\PassOptionsToPackage{unicode}{hyperref}
\PassOptionsToPackage{hyphens}{url}
\PassOptionsToPackage{dvipsnames,svgnames,x11names}{xcolor}
\documentclass[
  a4paper,
  11pt]{article}
\usepackage{xcolor}
\usepackage[margin=25mm]{geometry}
\usepackage{amsmath,amssymb}
\usepackage{iftex}
\ifPDFTeX
  \usepackage[T1]{fontenc}
  \usepackage[utf8]{inputenc}
  \usepackage{textcomp} 
\else 
  \usepackage{unicode-math} 
  \defaultfontfeatures{Scale=MatchLowercase}
  \defaultfontfeatures[\rmfamily]{Ligatures=TeX,Scale=1}
\fi
\usepackage{lmodern}
\ifPDFTeX\else
\fi
\IfFileExists{upquote.sty}{\usepackage{upquote}}{}
\IfFileExists{microtype.sty}{
  \usepackage[]{microtype}
  \UseMicrotypeSet[protrusion]{basicmath} 
}{}
\makeatletter
\@ifundefined{KOMAClassName}{
  \IfFileExists{parskip.sty}{%
    \usepackage{parskip}
  }{
    \setlength{\parindent}{0pt}
    \setlength{\parskip}{6pt plus 2pt minus 1pt}}
}{
  \KOMAoptions{parskip=half}}
\makeatother
\usepackage{color}
\usepackage{fancyvrb}

\DefineVerbatimEnvironment{Highlighting}{Verbatim}{commandchars=\\\{\}}
\newenvironment{Shaded}{}{}

\newcommand{\NormalTok}[1]{#1}

\usepackage{longtable,booktabs,array}
\usepackage{caption}
\usepackage{calc} 
\usepackage{etoolbox}
\makeatletter
\patchcmd\longtable{\par}{\if@noskipsec\mbox{}\fi\par}{}{}
\makeatother
\IfFileExists{footnotehyper.sty}{\usepackage{footnotehyper}}{\usepackage{footnote}}
\makesavenoteenv{longtable}
\usepackage{graphicx}
\makeatletter
\newsavebox\pandoc@box
\newcommand*\pandocbounded[1]{
  \sbox\pandoc@box{#1}%
  \Gscale@div\@tempa{\textheight}{\dimexpr\ht\pandoc@box+\dp\pandoc@box\relax}%
  \Gscale@div\@tempb{\linewidth}{\wd\pandoc@box}%
  \ifdim\@tempb\p@<\@tempa\p@\let\@tempa\@tempb\fi
  \ifdim\@tempa\p@<\p@\scalebox{\@tempa}{\usebox\pandoc@box}%
  \else\usebox{\pandoc@box}%
  \fi%
}
\def\fps@figure{htbp}
\makeatother
\providecommand{\tightlist}{%
  \setlength{\itemsep}{0pt}\setlength{\parskip}{0pt}}
\usepackage{microtype}
\usepackage[scale=0.88]{sourcecodepro}
\usepackage{longtable}
\usepackage{booktabs}
\usepackage{array}
\usepackage{colortbl}
\usepackage{titlesec}
\usepackage{titling}
\usepackage{fancyhdr}
\usepackage[most]{tcolorbox}
\usepackage{setspace}
\usepackage{enumitem}
\usepackage{ragged2e}
\usepackage{graphicx}
\graphicspath{{article/arxiv/}{./}}
\usepackage{hyperref}
\usepackage{pdflscape}
\usepackage{seqsplit}
\usepackage{ucharclasses}
\newfontfamily\tamilfont[
  Path=./,
  Script=Tamil,
  BoldFont=NotoSansTamil-Bold.ttf,
  Scale=0.96
]{NotoSansTamil-Regular.ttf}
\setTransitionsFor{Tamil}{\tamilfont}{\rmfamily}

\definecolor{MathdownPaper}{HTML}{FBFAF6}
\definecolor{MathdownInk}{HTML}{24231F}
\definecolor{MathdownMuted}{HTML}{78756D}
\definecolor{MathdownLine}{HTML}{D8D3C8}
\definecolor{MathdownAccent}{HTML}{B84F32}
\definecolor{MathdownAccentSoft}{HTML}{F1DFD8}
\definecolor{MathdownEditor}{HTML}{F3F0E8}
\pagecolor{MathdownPaper}
\color{MathdownInk}

\setlist{topsep=0.45em,itemsep=0.22em,parsep=0pt,leftmargin=1.6em}
\arrayrulecolor{MathdownLine}
\AtBeginEnvironment{longtable}{\arrayrulecolor{MathdownLine}}

\titleformat{\section}
  {\sffamily\bfseries\color{MathdownInk}\fontsize{18}{22}\selectfont}
  {}{0pt}{}
  [\vspace{0.15em}{\color{MathdownAccent}\titlerule[1.2pt]}]
\titleformat{\subsection}
  {\sffamily\bfseries\color{MathdownInk}\fontsize{14}{17}\selectfont}
  {}{0pt}{}
\titleformat{\subsubsection}
  {\sffamily\bfseries\color{MathdownInk}\fontsize{11.5}{14}\selectfont}
  {}{0pt}{}
\titlespacing*{\section}{0pt}{2.15em}{0.8em}
\titlespacing*{\subsection}{0pt}{1.65em}{0.45em}
\titlespacing*{\subsubsection}{0pt}{1.25em}{0.35em}

\pretitle{\begin{flushleft}\sffamily\bfseries\color{MathdownInk}%
  \fontsize{27}{31}\selectfont\raggedright}
\posttitle{\par\end{flushleft}\vspace{0.35em}%
  {\color{MathdownAccent}\rule{4.8em}{2.2pt}}\vspace{0.8em}}
\preauthor{\begin{flushleft}\sffamily\color{MathdownMuted}\small}
\postauthor{\par\end{flushleft}\vspace{-0.7em}}
\predate{\begin{flushleft}\sffamily\color{MathdownMuted}\footnotesize}
\postdate{\par\end{flushleft}\vspace{1.15em}}

\newtcolorbox{paperabstract}{
  enhanced,
  breakable,
  colback=MathdownAccentSoft,
  colframe=MathdownAccentSoft,
  boxrule=0pt,
  borderline west={2.4pt}{0pt}{MathdownAccent},
  arc=2.5mm,
  left=4.5mm,
  right=4.5mm,
  top=3.5mm,
  bottom=3.5mm,
  before skip=0.3em,
  after skip=1.4em
}
\newcommand{\paperabstracttitle}{%
  {\sffamily\bfseries\color{MathdownAccent}\large Abstract}\par\vspace{0.35em}}

\renewcommand{\headrulewidth}{0.4pt}
\renewcommand{\headrule}{\hbox to\headwidth{%
  \color{MathdownLine}\leaders\hrule height \headrulewidth\hfill}}
\AtBeginDocument{\RaggedRight}
\usepackage{bookmark}
\IfFileExists{xurl.sty}{\usepackage{xurl}}{} 
\makeatletter
\@ifundefined{xmpquote}{}{}
\makeatother
\hypersetup{
  pdftitle={Morphology-Aware Reversible Semantic Tokenization and Hierarchical Word Composition for Tamil Language Models},
  pdfauthor={Anand Murugan anand.murugan@gmail.com},
  colorlinks=true,
  linkcolor={MathdownAccent},
  filecolor={Maroon},
  citecolor={MathdownAccent},
  urlcolor={MathdownAccent},
  pdfcreator={LaTeX via pandoc}}

\title{Morphology-Aware Reversible Semantic Tokenization and
Hierarchical Word Composition for Tamil Language Models}
\author{Anand Murugan\\
\href{mailto:anand.murugan@gmail.com}{\nolinkurl{anand.murugan@gmail.com}}}
\date{2026-07-31}

\begin{document}
\maketitle

\begin{paperabstract}\paperabstracttitle

Statistical subword tokenizers efficiently split any input text into
tokens that a language model can process, but their units need not align
with lexical or grammatical structure. This is particularly
consequential for Tamil, where a written word can combine stem changes,
case, number, tense, agreement, voice, clitics and multiple linked
verbs. We present a \textbf{Tamil morphology system} built by extending
the open-source ThamizhiMorph analyzer and generator, a bounded
byte-exact semantic tokenizer and a learned hierarchical word composer.
\textbf{12 Finite-State Transducers (FSTs)} analyze Tamil words into
lemmas and grammatical features. The tokenizer can also reconstruct the
original text exactly, using Tamil character and byte fallbacks when
necessary.

The flat tokenizer (morphology-flat) represents each Tamil word with
lemmas and grammatical features as separate input tokens, exposing
useful semantic structure but producing significantly more tokens per
word. Our word-composer keeps the lemmas, but summarizes each word's
grammatical features into a single summary feature, and later uses
sentence context to recover the most relevant grammatical details for
the decoder.

We compare \textbf{morphology-flat}, the signal-preserving
\textbf{word-composer} and tokenizer arms based on \textbf{Sarvam-1,
AI4Bharat IndicBERTv2 and BrahmicTokenizer-131K}. All systems use the
same \textbf{69,591 Tamil--English training pairs},
\textbf{18.97-million-parameter encoder--decoder}, \textbf{40,000
updates}, English target tokenizer, optimizer, rotary positions,
numeric-copy policy and generation settings.

On the one-time \textbf{3,539-row} protected IN22/FLORES+ evaluation,
\textbf{morphology-flat achieves the best pooled scores: 10.63 BLEU,
35.26 chrF++ and 0.6276 COMETKiwi}. Compared with AI4Bharat---the
strongest external-tokenizer arm---these represent \textbf{relative
score improvements of 7.2\%, 3.2\% and 2.6\%}, respectively. The
word-composer scores \textbf{10.30, 34.88 and 0.6241, improving on
AI4Bharat by 3.8\%, 2.1\% and 2.0\%}, while also outperforming the
Sarvam and Brahmic arms.

Compared to morphology-flat, the composer reduces mean global source
states from \textbf{71.48 to 29.08} (by \textbf{59.3\%}). An
operation-count estimate gives the composer \textbf{9--21\% fewer
inference FLOPs}, depending on decoder caching. Its small quality
deficit is concentrated in longer formal FLORES+ sentences; the two IN22
partitions are paired ties. Later development-only tests show that
post-encoder retrieval of the original grammar factors is important,
while earlier retrieval and tested two-summary designs do not improve
the quality-efficiency tradeoff. These results show both the benefit and
the cost of explicit morphology. Lemmas and grammatical features improve
translation under the same small-model budget. The word-composer reduces
global attention states significantly and is estimated to require fewer
operations (FLOPs) overall, although it retains a small
translation-quality gap.

\end{paperabstract}

\section{1. Introduction}\label{1-introduction}

Subword tokenization made open-vocabulary neural machine translation
practical by representing rare words as reusable pieces rather than
fixed word IDs (\href{https://aclanthology.org/P16-1162/}{Sennrich et
al., 2016}). Language-independent implementations such as SentencePiece
can learn directly from raw text and avoid a language-specific
preprocessing pipeline (\href{https://aclanthology.org/D18-2012/}{Kudo
and Richardson, 2018}). These are strong engineering properties. They do
not, however, imply that a learned piece corresponds to a lemma, case
marker, tense, auxiliary relation or other linguistic category.

Tamil exposes this distinction clearly. The surface word \texttt{மரங்களை}
can be analyzed as \texttt{மரம்\ +\ noun\ +\ plural\ +\ accusative};
\texttt{படித்தான்} as
\texttt{படி\ +\ past\ +\ third-person\ singular\ masculine}; and
\texttt{படித்துக்கொடுத்தான்} as two lexical verbs connected by a typed
nonfinite relation. Stem changes prevent this structure from being
recovered by suffix removal alone: \texttt{மரம்\ →\ மரத்தை},
\texttt{ஆறு\ →\ ஆற்றை}, \texttt{காடு\ →\ காட்டை} and \texttt{பூ\ →\ பூவை}
follow different lexical classes.

This work asks two questions. First, can Tamil morphology be exposed to
a neural model in a deterministic, semantically explicit, auditable and
exactly reversible representation? Second, can a model exploit that
typically verbose representation without paying the full cost of flat
attention over every factor?

We separate the system into three layers:

\begin{enumerate}
\def\labelenumi{\arabic{enumi}.}
\tightlist
\item
  an FST-based morphology system, supported by lemma lexicons, that can
  analyze Tamil words into lemmas and grammatical features or generate
  words from those analyses;
\item
  a semantic tokenizer that produces a semantic token stream. It
  contains lemmas and grammatical features, along with any sandhi,
  spelling-variant, ambiguity-marker, structural, grapheme or byte
  tokens needed to reconstruct the original input exactly;
\item
  a neural interface that presents the same tokenized information to a
  language model either as a flat sequence (flat-morphology) or as a
  compact representation organized by word (word-composer).
\end{enumerate}

This separation keeps the tokenizer output easy to inspect while
allowing the model to reorganize the same information for efficiency.
Lemmas, grammatical features and reconstruction controls have different
purposes, but they remain in one exactly reversible token stream. This
design also gives us two clear comparisons. Comparing flat-morphology
with conventional statistical tokenizers tests the effect of changing
the source representation. Comparing the word-composer with
flat-morphology tests the model architecture while keeping the tokenizer
output exactly the same.

\pandocbounded{\includegraphics[keepaspectratio,alt={Flow from Tamil input through the readable semantic stream to the word composer}]{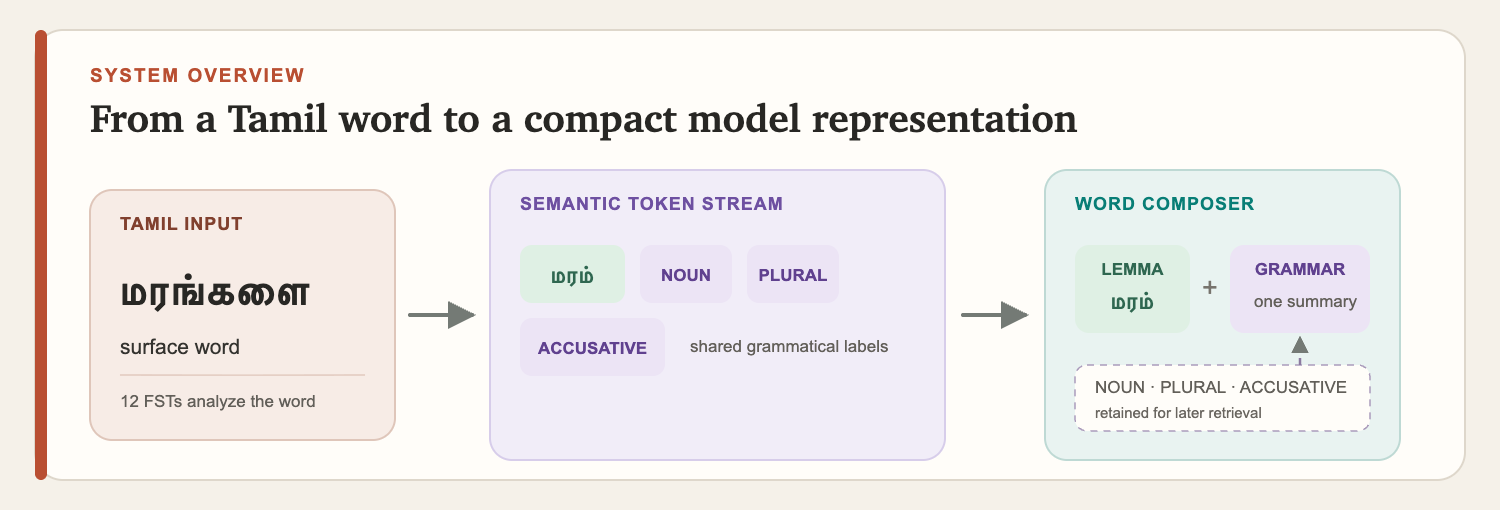}}

\textbf{Figure 1.} The tokenizer separates a Tamil word into its lemma
and grammatical features. The word-composer keeps the lemma and combines
the grammatical features into one summary for the model. It also keeps
the original features so the model can consult them later.

Morphology-flat produces the best protected-data translation quality in
one matched-parameter, one-seed small-model experiment. The
signal-preserving composer is second-best at protected-data translation
quality while reducing global token count by \textbf{59.3\%}. Both
morphology-flat and composer beat other external tokenizers. Later
development experiments explain several design choices: retrieving
grammar after sentence encoding helps; moving retrieval earlier hurts;
and neither a hand-written nor a learned two-summary design makes enough
use of its second state to justify the added positions.

Our contributions are:

\begin{itemize}
\tightlist
\item
  a Tamil morphology system built from 12 FST models, with clearly
  defined word classes, inflectional rules, compound forms, preservation
  of valid alternative analyses, and tests in both directions;
\item
  a semantic tokenizer with a fixed vocabulary that can reconstruct the
  original text exactly;
\item
  a fallback method that breaks unfamiliar words into Tamil character
  units before using bytes, so they do not collapse into an unknown
  token;
\item
  a controlled comparison between the flat-morphology token sequence and
  a word-composer adapter that shortens the model input without
  discarding information
\item
  tokenizer evaluations by training Tamil-to-English translation
  language models with comparison using the same data, model parameter
  count, training process, number handling and text-generation settings;
\item
  a reproducible release that records exact versions and file checksums
  for every important artifact.
\end{itemize}

\clearpage

\section{2. Tamil Morphological
System}\label{2-tamil-morphological-system}

\subsection{2.1 Construction and lexical
sources}\label{21-construction-and-lexical-sources}

The FST lineage begins with \textbf{ThamizhiMorph}, the open-source
Tamil morphological analyzer and generator implemented with Foma by
Sarveswaran, Dias and Butt
(\href{https://doi.org/10.1007/s10590-021-09261-5}{2021}). Its
meta-morph rule design was presented earlier by the same authors
(\href{https://aclanthology.org/W19-3111/}{2019}). The present work
builds directly on the
\href{https://github.com/sarves/thamizhi-morph}{released ThamizhiMorph
FST models and lexicons}, while substantially expanding and revising
lexical coverage, class structure, productive rules, ambiguity handling,
generation behavior and regression tests. In the original sources, nouns
were organized into 16 inflection classes and base verbs into 20 class
paths, counting classes 6.1, 6.2 and 6.3 separately. Our system has
\textbf{23 noun class or subclass paths} and \textbf{24 base or
irregular verb paths}, in addition to \textbf{70 productive auxiliary
continuations}. The corresponding explicit inventories grew from
\textbf{26,343 distinct noun roots to 127,311}, and from \textbf{2,850
distinct base-verb lexical strings to 11,399}. The released tokenizer
contains \textbf{139,899 lemma entries} across all lexical categories.
These extensions should not obscure the origin of the finite-state
implementation.

The expanded lexical inventory draws on three headword sources: the
\href{https://dsal.uchicago.edu/dictionaries/tamil-lex/}{University of
Madras \emph{Tamil Lexicon}}, digitized and hosted by the University of
Chicago\textquotesingle s Digital South Asia Library; official
\href{https://ta.wiktionary.org/wiki/முதற்_பக்கம்}{Tamil Wiktionary} title
and page dumps; and the supplemental
\href{https://github.com/Vuizur/Wiktionary-Dictionaries}{Vuizur
Wiktionary-Dictionaries} Tamil export. After normalization, the audited
snapshots contain \textbf{106,486 Tamil Lexicon lemmas}, \textbf{98,104
Tamil Wiktionary lemmas} and \textbf{5,509 Vuizur lemmas}. Because the
sources overlap, their deduplicated union is \textbf{129,396 lemmas},
with \textbf{76,518 attested in at least two sources}.

These word lists were used to expand the lexicon, define and refine
grammatical classes, and develop rules for generating inflected forms.
The release records the exact versions of these lexical resources,
verifies that the files have not changed, and documents their licenses
and redistribution restrictions.

\subsection{2.2 What the system contains and
generates}\label{22-what-the-system-contains-and-generates}

The system contains \textbf{12 FST models} for nouns, verbs, chained
verbs, adjectives, adverbs, pronouns, entities and smaller word classes.
The released model has \textbf{139,895 root lemmas}. Its two largest
groups are \textbf{127,311 noun roots} and \textbf{11,399 verb roots or
stems}. Some spellings occur in more than one group because the same
word can have both noun and verb readings.

Nouns are divided into \textbf{23 classes and subclasses} according to
how their stems change when endings are added. Verbs are divided into
\textbf{24 classes and subclasses}. The three largest parts of the
system have the following scale:

\begingroup\small\sffamily

{\def\LTcaptype{none} 
\begin{longtable}[]{@{}
  >{\raggedright\arraybackslash}p{(\linewidth - 6\tabcolsep) * \real{0.2000}}
  >{\raggedright\arraybackslash}p{(\linewidth - 6\tabcolsep) * \real{0.4200}}
  >{\raggedleft\arraybackslash}p{(\linewidth - 6\tabcolsep) * \real{0.1900}}
  >{\raggedleft\arraybackslash}p{(\linewidth - 6\tabcolsep) * \real{0.1900}}@{}}
\toprule\noalign{}
\begin{minipage}[b]{\linewidth}\raggedright
Part of the system
\end{minipage} & \begin{minipage}[b]{\linewidth}\raggedright
Lexical and rule inventory
\end{minipage} & \begin{minipage}[b]{\linewidth}\raggedleft
Grammatical analyses
\end{minipage} & \begin{minipage}[b]{\linewidth}\raggedleft
Written forms
\end{minipage} \\
\midrule\noalign{}
\endhead
\bottomrule\noalign{}
\endlastfoot
nouns & \textbf{127,311 roots} in \textbf{23 classes and subclasses} &
\textbf{70,014,090} & \textbf{66,340,679} \\
single verbs & \textbf{11,399 roots or stems} in \textbf{24 classes and
subclasses} & \textbf{164,917,397} & \textbf{152,589,463} \\
chained verbs & \textbf{20 following verbs} and \textbf{70 allowed ways
to continue a chain} & \textbf{2,456,063,902} &
\textbf{2,186,215,210} \\
\end{longtable}
}

\endgroup

The chained-verb figures are counts from that model by itself. They
include overlap with other models and intermediate forms used to build
longer words, so they cannot be added directly to the noun and
single-verb rows.

The noun vocabulary was also studied through \textbf{38 word-ending
families}. Long, specific endings were highly predictable:
\textbf{99.9\% of 2,969 nouns ending in \texttt{-னம்}} and \textbf{99.8\%
of 1,743 nouns ending in \texttt{-ியம்}} followed the expected pattern.
All \textbf{522 nouns ending in \texttt{-ிப்பு}}, all \textbf{218 ending
in \texttt{-ைப்பு}} and all \textbf{83 ending in \texttt{-வியல்}} also
followed their expected patterns. Shorter endings were less reliable:
only about \textbf{75\%} of words ending in \texttt{இ} and \textbf{73\%}
of words ending in short \texttt{உ} followed the expected noun pattern.
This shows why a word\textquotesingle s ending is useful evidence but
cannot by itself determine the correct class.

Tamil can also combine several verbs inside one written word. The system
uses \textbf{20 common auxiliary or light verbs} and \textbf{70
continuation patterns}. These patterns are stored in \textbf{47 groups},
allowing verbs that support the same combinations to share rules. Before
tense and person endings are added, the rules allow \textbf{4,913,950
combinations of a starting verb, a connector and a following verb}.

After overlap and intermediate-only forms are removed across the
complete system, chained verbs still account for more than \textbf{1.9
billion} forms, while nouns, single verbs and the other basic models
together account for about \textbf{205.5 million}.

The smaller models are also substantial:

\begingroup\small\sffamily

{\def\LTcaptype{none} 
\begin{longtable}[]{@{}
  >{\raggedright\arraybackslash}p{(\linewidth - 6\tabcolsep) * \real{0.3000}}
  >{\raggedleft\arraybackslash}p{(\linewidth - 6\tabcolsep) * \real{0.2200}}
  >{\raggedleft\arraybackslash}p{(\linewidth - 6\tabcolsep) * \real{0.2400}}
  >{\raggedleft\arraybackslash}p{(\linewidth - 6\tabcolsep) * \real{0.2400}}@{}}
\toprule\noalign{}
\begin{minipage}[b]{\linewidth}\raggedright
Model
\end{minipage} & \begin{minipage}[b]{\linewidth}\raggedleft
Base words
\end{minipage} & \begin{minipage}[b]{\linewidth}\raggedleft
Grammatical analyses
\end{minipage} & \begin{minipage}[b]{\linewidth}\raggedleft
Written forms
\end{minipage} \\
\midrule\noalign{}
\endhead
\bottomrule\noalign{}
\endlastfoot
standalone adjectives & \textbf{662} & \textbf{3,081} &
\textbf{3,080} \\
adverbs & \textbf{932} & \textbf{4,592} & \textbf{4,584} \\
pronouns & \textbf{44} & \textbf{1,723} & \textbf{1,594} \\
particles and other function words & \textbf{414} & \textbf{2,568} &
\textbf{2,471} \\
\end{longtable}
}

\endgroup

The adjective and noun models divide their work to avoid duplication.
The adjective model handles words that are directly classified as
adjectives and forms built from adjective bases. Adjective-like forms
built from nouns are handled by the noun model because it already knows
how each noun stem changes. \textbf{17 singular and 20 plural noun
patterns} generate forms ending in \texttt{-ஆன} and \texttt{-அற்ற},
together with related noun forms. These add about \textbf{1,011,524
valid modifier analyses} outside the standalone adjective model.
Removing the same noun-based rules from the adjective model eliminated
duplicate analyses without losing any generated forms.

The adverb model is similarly careful. It contains \textbf{771 words
with direct adverb readings} and \textbf{932 base words} when words that
can also act as nouns, adjectives, pronouns or verbs are included. It
does not classify every word ending in \texttt{-ஆக} as an adverb. Some
such words describe a role or purpose, while others belong to verb
constructions. Preserving these differences avoids incorrect analyses,
even if more unsupported words must use the fallback system.

The \textbf{129,020-entry source lemma dictionary} is a separate
coverage checklist, not the complete model vocabulary. The model
vocabulary also contains reviewed roots from its grammatical-class
inventories and lemmas needed to represent compound words as known
parts.

\begingroup\small\sffamily

{\def\LTcaptype{none} 
\begin{longtable}[]{@{}
  >{\raggedright\arraybackslash}p{(\linewidth - 2\tabcolsep) * \real{0.5800}}
  >{\raggedleft\arraybackslash}p{(\linewidth - 2\tabcolsep) * \real{0.4200}}@{}}
\toprule\noalign{}
\begin{minipage}[b]{\linewidth}\raggedright
Source-dictionary coverage
\end{minipage} & \begin{minipage}[b]{\linewidth}\raggedleft
Lemmas
\end{minipage} \\
\midrule\noalign{}
\endhead
\bottomrule\noalign{}
\endlastfoot
directly recognized by at least one FST & \textbf{102,726} \\
not directly recognized by any FST & \textbf{26,294} \\
total source-dictionary checklist & \textbf{129,020} \\
\end{longtable}
}

\endgroup

The unrecognized group of \textbf{26,294 lemmas} is mostly historical or
rare material, names, noisy dictionary entries and words whose
grammatical class remains uncertain. It may also contain genuine
coverage gaps. We do not automatically place every unmatched word into a
noun or verb class, because one unsupported class assignment could
generate a large family of incorrect forms. Such input is still
preserved exactly through the tokenizer\textquotesingle s grapheme and
byte fallback, but it does not receive an invented morphological
analysis.

After duplicate spellings across the models are counted only once, the
complete system accepts \textbf{2,130,878,180 distinct written forms}
and \textbf{2,107,907,215 grammatical analyses}. This count excludes
forms that exist only as intermediate steps for joining another ending.
It does not mean that Tamil has \textbf{2.13 billion} ordinary
dictionary words. It shows how fewer than \textbf{140,000 root lemmas}
can produce billions of inflected and multi-verb forms.

The same spelling can have more than one analysis, and one analysis can
sometimes allow more than one spelling:

\begin{itemize}
\tightlist
\item
  \textbf{One spelling, two analyses:} \texttt{மரத்தால்} can be analyzed
  as the noun \texttt{மரம்} with the meaning ``by or with the tree,'' or
  as a conditional verb form of \texttt{மர}, depending on the sentence.
\item
  \textbf{One analysis, two spellings:} the analysis
  \texttt{சுவர்\ +\ noun\ +\ accusative} can be written as either
  \texttt{சுவரை} or \texttt{சுவற்றை}.
\end{itemize}

When the lemma is ignored, the system uses \textbf{36,058 grammatical
patterns} after final Sandhi (ஒற்றெழுத்து) sound-linking markers are
removed, or \textbf{46,846} when those markers are retained. Counting
the same pattern separately in each model gives \textbf{54,263
model-specific patterns}. Exact reconstruction requires \textbf{94,569
spelling-sensitive patterns}. These compact pattern inventories allow
the tokenizer to represent the full system without assigning a separate
vocabulary entry to every generated word.

\subsection{2.3 Ambiguity and
validation}\label{23-ambiguity-and-validation}

The analyzer preserves distinct valid readings. A deterministic
context-free ranker supplies a default where needed, but candidate
analyses remain available. We treat analysis recall, invalid-analysis
rate and best-reading accuracy as separate quantities.

Validation includes FST build regressions, forward and inverse probes,
paradigm generation, exact-pattern witnesses and source-lemma audits.
The current release reports \textbf{574,379 regenerated paradigm probes
with zero failures} and \textbf{94,569 exact realization witnesses with
zero round-trip failures}. These are structural tests, not a claim of
complete linguistic accuracy.

\subsection{2.4 Development corpus
audits}\label{24-development-corpus-audits}

Several corpora were also inspected while developing and auditing
morphology coverage. The general-text audit used the full
\href{https://huggingface.co/datasets/mozhi-ai/tamil-corpus}{Mozhi Tamil
corpus}, two pinned samples of the verified Tamil portion of
\href{https://huggingface.co/datasets/ai4bharat/sangraha}{Sangraha}
(\href{https://arxiv.org/abs/2403.06350}{Khan et al., 2024}), the
training splits of
\href{https://huggingface.co/datasets/IsaacRodgz/DravidianCodeMix-Dataset}{Dravidian
CodeMix} and
\href{https://huggingface.co/datasets/dheepakkaran/TamilTech-QA}{TamilTech-QA}.
The train and development splits of
\href{https://universaldependencies.org/treebanks/ta_ttb/index.html}{UD
Tamil-TTB}
(\href{http://www.lrec-conf.org/proceedings/lrec2012/summaries/456.html}{Ramasamy
and Žabokrtský, 2012}) supported a separate lemma, part-of-speech and
feature diagnostic, while the Tamil training split of
\href{https://huggingface.co/datasets/ai4bharat/naamapadam}{Naamapadam}
(\href{https://aclanthology.org/2023.acl-long.582/}{Mhaske et al.,
2023}) was used to stress-test entity handling. These resources exposed
coverage gaps and supplied regression candidates. Because they were
inspected during development, none is treated as held-out evidence of
tokenizer or translation quality.

\section{3. Reversible Semantic
Tokenization}\label{3-reversible-semantic-tokenization}

BPE-based tokenizers represent rare words through subword units learned
from frequency statistics
(\href{https://aclanthology.org/P16-1162/}{Sennrich et al., 2016}).
SentencePiece supports BPE and unigram tokenization directly from raw
text (\href{https://aclanthology.org/D18-2012/}{Kudo and Richardson,
2018}). These approaches are general, fast and widely supported.

Morphology-aware and factored tokenizers instead supply a model with
information such as lemma, part of speech and inflectional features. Our
representation differs from a lossy morphological segmenter in two
respects. It explicitly identifies lexical and grammatical features
across different word classes, rather than only marking morphological
boundaries, and its public codec can reconstruct the exact original
UTF-8 input.

\subsection{3.1 Semantic stream}\label{31-semantic-stream}

An analysis is mapped deterministically to lemma and feature tokens. For
example:

\begin{Shaded}
\begin{Highlighting}[]
\NormalTok{மரங்களை}
\NormalTok{மரம் \textless{}POS\_NOUN\textgreater{} \textless{}NUM\_PL\textgreater{} \textless{}CASE\_ACC\textgreater{}}

\NormalTok{படித்துக்கொடுத்தான்}
\NormalTok{படி \textless{}LINK\_VPART\textgreater{} கொடு \textless{}TENSE\_PAST\textgreater{} \textless{}PERSON\_3SG\_MASC\textgreater{}}
\end{Highlighting}
\end{Shaded}

The same grammatical tokens are shared across words even when Tamil
applies different spelling rules. The following nouns all have
accusative case, but their stems change in different ways when the
ending is added:

\begingroup\small\sffamily

{\def\LTcaptype{none} 
\begin{longtable}[]{@{}
  >{\raggedright\arraybackslash}p{(\linewidth - 4\tabcolsep) * \real{0.1500}}
  >{\raggedright\arraybackslash}p{(\linewidth - 4\tabcolsep) * \real{0.5000}}
  >{\raggedright\arraybackslash}p{(\linewidth - 4\tabcolsep) * \real{0.3500}}@{}}
\toprule\noalign{}
\begin{minipage}[b]{\linewidth}\raggedright
Written word
\end{minipage} & \begin{minipage}[b]{\linewidth}\raggedright
Semantic token output
\end{minipage} & \begin{minipage}[b]{\linewidth}\raggedright
Visible spelling change
\end{minipage} \\
\midrule\noalign{}
\endhead
\bottomrule\noalign{}
\endlastfoot
\texttt{பூவை} &
\texttt{பூ\ \textless{}POS\_NOUN\textgreater{}\ \textless{}CASE\_ACC\textgreater{}}
& \texttt{வ்} is inserted before the ending \\
\texttt{காட்டை} &
\texttt{காடு\ \textless{}POS\_NOUN\textgreater{}\ \textless{}CASE\_ACC\textgreater{}}
& the final \texttt{உ} disappears and \texttt{ட} doubles \\
\texttt{வண்டை} &
\texttt{வண்டு\ \textless{}POS\_NOUN\textgreater{}\ \textless{}CASE\_ACC\textgreater{}}
& the final \texttt{உ} disappears without doubling \\
\texttt{ஆற்றை} &
\texttt{ஆறு\ \textless{}POS\_NOUN\textgreater{}\ \textless{}CASE\_ACC\textgreater{}}
& final \texttt{று} changes to \texttt{ற்று} \\
\texttt{மரத்தை} &
\texttt{மரம்\ \textless{}POS\_NOUN\textgreater{}\ \textless{}STEM\_OBLIQUE\textgreater{}\ \textless{}CASE\_ACC\textgreater{}}
& final \texttt{ம்} changes to \texttt{த்து} \\
\end{longtable}
}

\endgroup

The same principle applies to verbs. These words have different written
past stems, but they share the tokens for past tense and third-person
singular masculine agreement:

\begin{Shaded}
\begin{Highlighting}[]
\NormalTok{படித்தான்    படி  \textless{}TENSE\_PAST\textgreater{} \textless{}PERSON\_3SG\_MASC\textgreater{}}
\NormalTok{விட்டான்     விடு \textless{}TENSE\_PAST\textgreater{} \textless{}PERSON\_3SG\_MASC\textgreater{}}
\NormalTok{பெற்றான்     பெறு \textless{}TENSE\_PAST\textgreater{} \textless{}PERSON\_3SG\_MASC\textgreater{}}
\NormalTok{சென்றான்     செல் \textless{}TENSE\_PAST\textgreater{} \textless{}PERSON\_3SG\_MASC\textgreater{}}
\NormalTok{கொண்டான்     கொள் \textless{}TENSE\_PAST\textgreater{} \textless{}PERSON\_3SG\_MASC\textgreater{}}
\end{Highlighting}
\end{Shaded}

The lemma identifies the word, while the shared tokens state its
grammatical meaning. The FST knows the spelling rules for each
lemma\textquotesingle s class, so the semantic stream does not need a
different past-tense or accusative token for every written stem change.
The exact codec described below records any extra choice needed when the
same analysis permits more than one valid spelling.

The fixed vocabulary contains \textbf{140,922 tokens}, including
\textbf{222 grammatical and semantic labels}, \textbf{139,957 lexical
tokens}, \textbf{378 Tamil grapheme fallback tokens} and \textbf{256
UTF-8 byte tokens}. Among the lexical tokens, \textbf{58} are used as
secondary lemmas in current multi-lemma and auxiliary analyses. They
remain lexical states rather than forming a special semantic token
class. Typed links distinguish verbal-participle, infinitive and other
compound relations instead of replacing them with a generic auxiliary
marker.

\subsection{3.2 Exact realization}\label{32-exact-realization}

The tokenizer first chooses a morphological analysis for the input word.
The FST can usually turn that analysis back into the expected spelling.
When the same analysis allows more than one valid spelling, the
tokenizer records a small marker that tells the decoder which spelling
appeared in the original text. Spaces, punctuation and text that the FST
does not recognize are also recorded directly. The goal is always

\[
\operatorname{decode}(\operatorname{encode}(x)) = x
\]

for any UTF-8 text accepted by the tokenizer. The decoder must be able
to restore the text from the tokens alone, without looking at the
original input.

The fallback hierarchy is:

\begin{enumerate}
\def\labelenumi{\arabic{enumi}.}
\tightlist
\item
  direct FST analysis;
\item
  reviewed entity analysis;
\item
  constrained semantic paths where enabled;
\item
  fixed Tamil grapheme units;
\item
  UTF-8 bytes.
\end{enumerate}

Fallback and unknown IDs are not equivalent. A grapheme or byte sequence
can retain the complete input even when no morphology analysis exists.

\subsection{3.3 How morphology coverage affects the token
stream}\label{33-how-morphology-coverage-affects-the-token-stream}

The tokenizer can provide lemma and grammatical tokens only when an FST
recognizes the word. When it does not, the fallback system still
preserves the original spelling exactly, but it usually needs more token
positions and provides less direct grammatical information. The opening
line of the Thirukkural gives a compact example of this difference.

An early release recognized only \textbf{four of its seven words}. It
represented the line with \textbf{95 tokens}, including \textbf{75 byte
tokens} used mainly to preserve the unrecognized text. A wider coverage
review found missing words and rules that also affected other Tamil
text. After those problems were fixed, the FSTs recognized \textbf{all
seven words}. The exact stream fell to \textbf{41 tokens}, and its
\textbf{six remaining byte tokens} represented spaces rather than
unrecognized Tamil words.

This example shows why FST coverage affects both the information in the
stream and its length. It also shows a limit of the approach: even after
repair, the semantic stream was longer than the \textbf{nine tokens}
produced by a compact BPE tokenizer. One sentence cannot establish
overall efficiency, so the later experiments measure sequence length
across the complete evaluation data.

\subsection{3.4 Public and model-facing
views}\label{34-public-and-model-facing-views}

The project distinguishes:

\begin{itemize}
\tightlist
\item
  human-readable semantic tokens;
\item
  the exact reversible codec;
\item
  diagnostic analysis records;
\item
  the translation language model adapter;
\item
  compact model-internal composer states.
\end{itemize}

Model-only compaction removes standalone \texttt{WORD\_START} key/value
positions and moves reconstruction-only \texttt{SURFACE\_BYTES} markers
into typed metadata. It elides a generic feature only when a more
specific feature provably implies it. POS and semantically informative
deictic information remain explicit. These optimizations do not alter
the public readable representation.

\section{4. Hierarchical word
composition}\label{4-hierarchical-word-composition}

The Transformer performs context-dependent attention over a sequence
(\href{https://arxiv.org/abs/1706.03762}{Vaswani et al., 2017}), and
successive layers can learn hierarchical patterns.

Our flat morphology stream expands each source word into a lemma and
multiple grammatical tokens. This lengthens the model-facing input and
makes sentence-level global attention more expensive. Our word-composer
compensates for this verbosity by retaining lexical states directly
while summarizing grammatical information. After the encoder has
processed the sentence, the composer uses the contextualized summary to
retrieve relevant information from the original fine-grained grammatical
factors before passing the resulting representation to the decoder.

\pandocbounded{\includegraphics[keepaspectratio,alt={Tamil semantic tokenizer and word-composer translation architecture}]{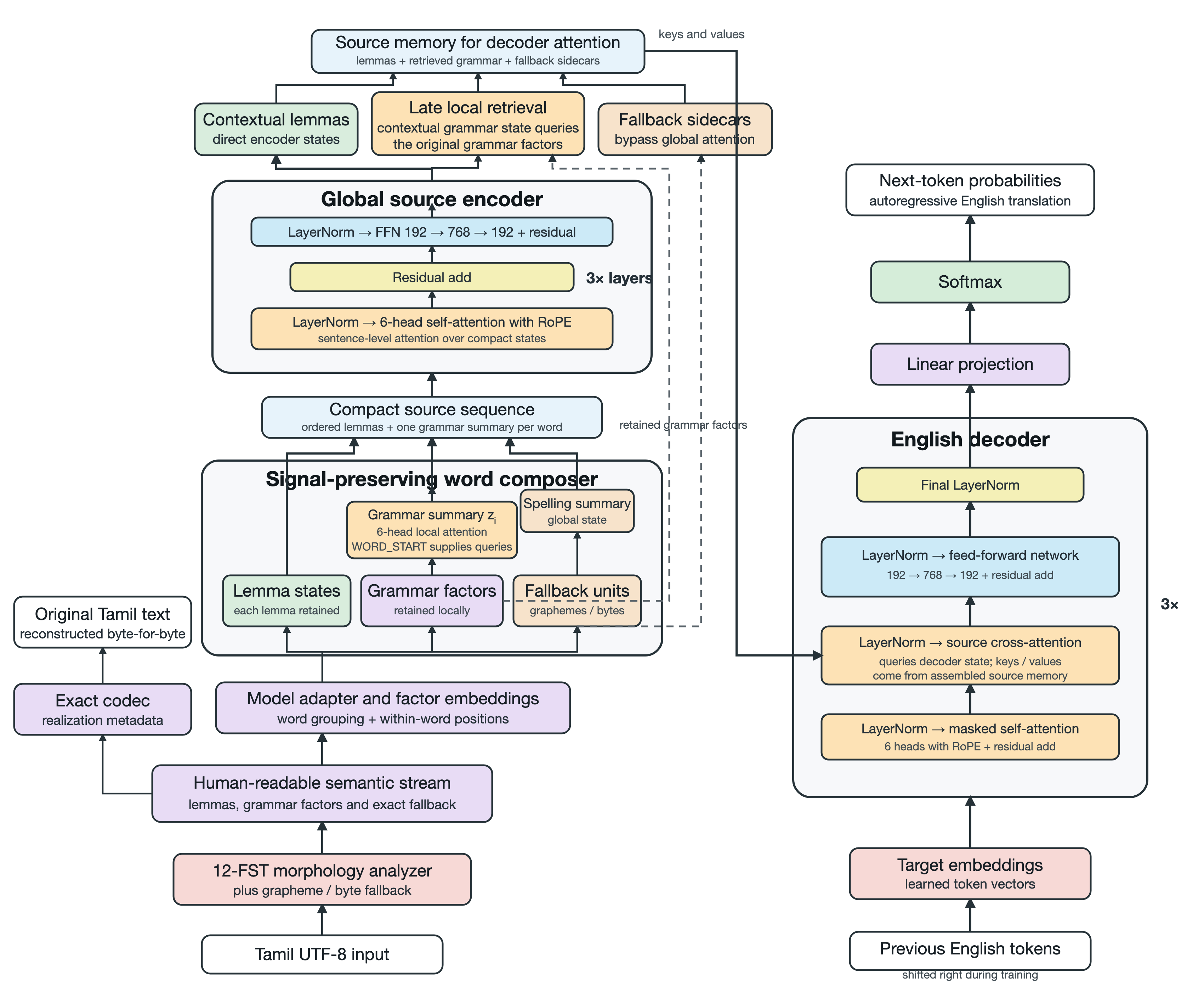}}

\textbf{Figure 2.} The tokenizer keeps a public, human-readable stream
that can reconstruct the Tamil input exactly. The model adapter
reorganizes the same information for translation: lemmas remain direct
sentence-level states, grammar is summarized and later retrieved from
the original factors, and fallback spelling units remain directly
available to the decoder.

\subsection{4.1 Signal-preserving
composer}\label{41-signal-preserving-composer}

For an analyzed word \(i\), let \(x_{ij}\) be the vector for its \(j\)th
tokenizer factor and let \(F_i\) be the number of factors in that word.
The flat model passes every factor through sentence-level attention, so
the sentence has

\[
N=\sum_i F_i
\]

source positions, ignoring the beginning and end markers for simplicity.

The composer divides the factor indices into two sets. \(L_i\) contains
the lemma positions, and \(G_i\) contains the grammatical-factor
positions. Every lemma vector \(x_{ij}\) for which \(j\in L_i\) is kept
as a separate sentence-level state. A chained-verb analysis can
therefore retain several lemmas in their original order.

The grammatical factors are replaced at the sentence level by one
summary vector. No grammar summary exists yet when this initial local
attention is performed. To create it, the model uses the learned
\texttt{\textless{}WORD\_START\textgreater{}} embedding, together with
its within-word position information, as the seed vector \(s_i\). Each
of the six attention heads turns this seed into one query. The grammar
factors of the same word provide the keys and values:

\[
\begin{aligned}
q_i^h &= W_Q^h s_i,\\
k_{ij}^h &= W_K^h x_{ij}, \quad j\in G_i,\\
v_{ij}^h &= W_V^h x_{ij}, \quad j\in G_i,\\
a_{ij}^h &= \operatorname{softmax}_j
\left(\frac{q_i^h\cdot k_{ij}^h}{\sqrt{d_h}}\right),\\
z_{i,\mathrm{gram}}^h &= \sum_{j\in G_i} a_{ij}^h v_{ij}^h.
\end{aligned}
\]

Here \(h\) identifies the attention head; \(q_i^h\) is that
head\textquotesingle s query for word \(i\); and \(k_{ij}^h\) and
\(v_{ij}^h\) are the key and value made from grammar factor \(j\). The
weight \(a_{ij}^h\) measures how much that factor contributes to the
head\textquotesingle s summary \(z_{i,\mathrm{gram}}^h\). The model
computes these scores only between the seed-derived query and
grammatical factors from the same word. It does not compare the factors
with one another or with factors from other words at this stage.

The six head summaries are joined and passed through an output
projection, residual connection, normalization and feed-forward network
to produce one grammar state \(z_i\). With \(d_{\text{model}}=192\),
each of the six heads has \(d_h=32\) channels. The compact
representation of an analyzed word is therefore

\[
C_i = \left(x_{ij}:j\in L_i\right)\mathbin{\|}z_i,
\]

where \(\mathbin{\|}\) means that the retained lemma states and grammar
summary are placed next to one another in the output sequence. More
precisely, the word-composer places all lemma states first, in their
original order, and then places one grammar summary for the entire word.
It does not alternate each lemma with a separate summary.

For example, the semantic stream for படித்துக்கொடுத்தான்,

\begin{Shaded}
\begin{Highlighting}[]
\NormalTok{படி \textless{}LINK\_VPART\textgreater{} கொடு \textless{}TENSE\_PAST\textgreater{} \textless{}PERSON\_3SG\_MASC\textgreater{}}
\end{Highlighting}
\end{Shaded}

becomes two direct lemma states followed by one grammar state:

\begin{Shaded}
\begin{Highlighting}[]
\NormalTok{[படி] [கொடு] [z\_i]}
\end{Highlighting}
\end{Shaded}

Here \(z_i\) summarizes \texttt{\textless{}LINK\_VPART\textgreater{}},
\texttt{\textless{}TENSE\_PAST\textgreater{}} and
\texttt{\textless{}PERSON\_3SG\_MASC\textgreater{}}. The original
within-word position embeddings remain part of the lemma and
grammatical-factor vectors, allowing the local attention to distinguish
their original locations.

All three global encoder layers then process the compact sentence. The
lemma states exchange information with the rest of the sentence without
first being compressed into the grammar summary. The encoder also adds
sentence context to the grammar-summary position. Only at this later
stage does the contextualized grammar summary become a query. It queries
the original fine-grained grammar factors of its own word, replacing the
earlier summary with a context-sensitive one for the decoder. This late
retrieval allows sentence context to affect which original factors
receive the most attention, without restoring every factor as a global
source position.

The initial summary and late retrieval use the same local attention and
feed-forward parameters, which are also shared with the first global
encoder layer. Local dropout is zero. The matched global encoder and
decoder dropout remain 0.1. This is not additive feature packing: the
original grammar factors remain separate vectors during both local
attention operations.

Fallback words follow a separate path. One spelling summary participates
in global sentence attention, while the original grapheme or byte
vectors remain available directly to decoder cross-attention. They are
not treated as grammatical factors or forced through the grammar
summary.

If \(P\) is the number of retained lemma states, grammar summaries,
fallback summaries and sentence markers, then \(P<N\) for the compact
sequence. The leading attention cost changes from
\(O\left(\left(\sum_i F_i\right)^2\right)\) for flat sentence attention
to \(O\left(\sum_i F_i\right)+O(P^2)\) for local composition and
retrieval plus global attention. This simplified comparison leaves out
projections, feed-forward networks, padding, decoder cross-attention and
decoding.

\subsection{4.2 Estimated computation}\label{42-estimated-computation}

Wall-clock measurements on a shared laptop are sensitive to background
load, thermal throttling and other activity. We therefore use an
approximate floating-point operation count to compare the two
architectures. The estimate uses unpadded examples at the recorded mean
training lengths: \textbf{80.77 flat source positions}, \textbf{32.44
composer global states}, \textbf{52.74 word-composer decoder-memory
states} and \textbf{29.55 target tokens}. It includes the
word-composer\textquotesingle s initial word-local attention and late
retrieval. One multiplication and one addition are counted as two
floating-point operations.

\begingroup\small\sffamily

{\def\LTcaptype{none} 
\begin{longtable}[]{@{}
  >{\raggedright\arraybackslash}p{(\linewidth - 6\tabcolsep) * \real{0.4900}}
  >{\raggedleft\arraybackslash}p{(\linewidth - 6\tabcolsep) * \real{0.1700}}
  >{\raggedleft\arraybackslash}p{(\linewidth - 6\tabcolsep) * \real{0.1700}}
  >{\raggedleft\arraybackslash}p{(\linewidth - 6\tabcolsep) * \real{0.1700}}@{}}
\toprule\noalign{}
\begin{minipage}[b]{\linewidth}\raggedright
Estimated work per average example
\end{minipage} & \begin{minipage}[b]{\linewidth}\raggedleft
Morphology-flat
\end{minipage} & \begin{minipage}[b]{\linewidth}\raggedleft
Composer
\end{minipage} & \begin{minipage}[b]{\linewidth}\raggedleft
Composer reduction
\end{minipage} \\
\midrule\noalign{}
\endhead
\bottomrule\noalign{}
\endlastfoot
source encoder, forward pass & 0.229 GFLOPs & 0.130 GFLOPs &
\textbf{43\%} \\
complete training forward pass & 0.546 GFLOPs & 0.432 GFLOPs &
\textbf{21\%} \\
training forward and backward, approximate & 1.64 GFLOPs & 1.30 GFLOPs &
\textbf{21\%} \\
30-token generation with the current uncached decoder & 5.71 GFLOPs &
5.21 GFLOPs & \textbf{9\%} \\
30-token generation with decoder key/value caching & 0.55 GFLOPs & 0.44
GFLOPs & \textbf{21\%} \\
\end{longtable}
}

\endgroup

The estimate suggests that the composer requires less arithmetic
overall, despite performing two word-local attention operations. Its
largest saving is in the source encoder, where fewer states participate
in sentence-level attention.

Fewer operations do not automatically produce shorter wall-clock time.
The composer currently uses many small variable-length operations,
indexing and memory copies, while the flat model uses larger regular
matrix multiplications that hardware libraries execute efficiently. The
present decoder also recomputes the complete English prefix at every
generation step. The FLOP counts therefore describe the
architecture\textquotesingle s computational potential, not a measured
latency guarantee. A controlled benchmark and a fused local composer
implementation are needed before making claims about end-to-end speed.

\subsection{4.3 What the architecture tests taught
us}\label{43-what-the-architecture-tests-taught-us}

The final composer and model architecture was not chosen from intuition
alone. We tested several ways of shortening or enriching the word
representation.

An earlier prototype compressed every word into one content-independent
summary and replaced one global encoder layer. It trained faster, but on
the \textbf{69,591-pair run} it lost \textbf{1.03 BLEU, 1.56 chrF++ and
0.0152 COMETKiwi} relative to flat morphology. This failure motivated
composer\textquotesingle s direct lemma path, separate grammar state,
late retrieval and full encoder depth.

We then trained matched versions of the composer with and without late
retrieval. Removing retrieval increased development cross-entropy by
\textbf{1.93\%} and reduced chrF++ by \textbf{1.58\%}. Directly
bypassing retrieval in trained models raised loss by about
\textbf{17-22\%}, depending on the diagnostic and checkpoint. The model
therefore uses the second look at the original grammar factors; it is
not an unused extra operation.

Moving retrieval earlier, between encoder layers two and three, did not
help. At \textbf{20,000 updates}, this model increased cross-entropy
from \textbf{2.6793 to 2.7180} and reduced chrF++ from about
\textbf{45.55 to 45.20}. The final encoder layer appears to mix away
some of the newly retrieved detail, so our current composer instead
retrieves immediately before decoder access.

We also tested two grammar summaries. We used a hand-written split
between factor groups. It preserved the assigned information, but the
second summary was sparse and weakly used: only \textbf{8.92\% of source
words} received it, \textbf{44.91\%} of those summaries contained one
factor, and removing the second state raised causal loss by only
\textbf{0.57\%}. Its small \textbf{40,000-update} development-score lead
did not come with a demonstrated subgroup benefit, and mean global
states rose from \textbf{32.44 to 33.80}.

A second experiment let two learned queries inspect every grammar factor
instead of imposing a hand-written split. The two states became
increasingly similar. On a measure where 1 means that two vectors point
in the same direction, their mean similarity reached \textbf{0.929}
after late retrieval. Removing the second state raised loss by only
\textbf{0.54\%}, while using two states increased the global source
sequence by \textbf{37.8\%} on the diagnostic set. This design was
stopped at \textbf{20,000 updates}.

These negative results narrow the claim. One summary is not guaranteed
to be perfect, but the tested second summaries added positions without
showing enough distinct, useful work. Our proposed composer is therefore
the best supported compact architecture in this study.

\section{5. Data and experimental
design}\label{5-data-and-experimental-design}

\subsection{5.1 Training data}\label{51-training-data}

We first tested our data-cleaning and review process on fixed samples of
\textbf{1,000 and 100,000} Tamil--English sentence pairs from
\textbf{Samanantar} (Ramesh et al., 2022). This testing revealed
inconsistent text formatting, repeated English translations, number
mismatches and incorrectly aligned sentence pairs. We therefore did not
include any Samanantar pairs in the final \textbf{69,591-pair} training
dataset.

The final training union contains \textbf{69,591 Tamil--English pairs}.
Its \textbf{22,955 authentic pairs} comprise \textbf{18,000 BPCC-Human
Wiki pairs}, \textbf{4,000 BPCC-Human Daily pairs} (Gala et al., 2023),
and \textbf{955 sentence pairs} aligned from corresponding Tamil and
English \href{https://www.pib.gov.in/}{Press Information Bureau}
releases. The remaining \textbf{46,636 pairs} begin with known-original
English sentences from two non-overlapping tranches of WikiText-103
(Merity et al., 2017). The pinned \textbf{IndicTrans2 English-to-Tamil
model} (Gala et al., 2023) generates their Tamil sources. The first
tranche contributed \textbf{19,409 reviewed pairs} to the earlier
\textbf{42,364-pair treatment}. For the second tranche, bidirectional
COMETKiwi and structural checks filtered \textbf{35,000 candidates to
27,234}.

A deterministic blinded Codex review sampled \textbf{200 rows}:
\textbf{194 were rated good} and six partial. The six sampled partials
and one exact prior-source collision were removed. The sample estimates
the gate's selectivity; it neither proves every unreviewed row correct
nor constitutes bilingual human evaluation. Synthetic Tamil may
reproduce the teacher's constructions and biases.

The authentic \textbf{2,000-row BPCC-Human Wiki development set} is
excluded from training. Exact source, target and pair overlap checks are
zero.

\subsection{5.2 Protected evaluation}\label{52-protected-evaluation}

The final test contains \textbf{1,024 IN22-Gen}, \textbf{1,503
IN22-Conv} and \textbf{1,012 FLORES+ devtest rows}. IN22-Gen and
IN22-Conv were introduced with IndicTrans2 and BPCC (Gala et al., 2023).
FLORES+ derives from the multilingual FLORES evaluation lineage;
FLORES-style sets use professionally translated parallel sentences at
fixed split sizes (\href{https://arxiv.org/abs/2207.04672}{NLLB Team et
al., 2022}). The exact dataset revisions and files are checksum-pinned.

The \textbf{3,539 rows} were opened once after checkpoints, tokenizers,
decoder settings and analysis procedures were frozen. Dataset
fingerprints, row order, checkpoint hashes, canonical tokenizer/FST
provenance, zero truncation and bit-exact checkpoint reloads were
verified. FLORES+ \texttt{dev} remained unopened.

\subsection{5.3 Arms and controls}\label{53-arms-and-controls}

The five primary systems are \textbf{morphology-flat}, the
signal-preserving \textbf{word-composer}, \textbf{Sarvam-1, AI4Bharat
IndicBERTv2 and BrahmicTokenizer-131K}. The frozen evaluation also
retains the earlier single-summary composer as a historical architecture
ablation.

The external comparison set uses the tokenizer artifacts from Sarvam-1,
AI4Bharat IndicBERTv2 and BrahmicTokenizer-131K. These artifacts are
used only as source tokenizers inside our randomly initialized
translation model. Our results do not evaluate the
organizations\textquotesingle{} complete pretrained systems. IndicBERTv2
is described with the IndicCorp and IndicXTREME resources by
\href{https://aclanthology.org/2023.acl-long.693/}{Doddapaneni et al.
(2023)}.

Every system uses \textbf{18,967,296 parameters}, \textbf{width 192},
\textbf{6 attention heads}, \textbf{FFN width 768}, \textbf{3 decoder
layers}, a \textbf{16,000-token English target tokenizer}, AdamW at
\textbf{learning rate 0.0007}, \textbf{batch size 24}, \textbf{40,000
updates} and seed 20260731. \textbf{RoPE} is applied to encoder and
decoder self-attention; cross-attention remains unrotated. Training
rows, order, target IDs, optimizer, schedule, numeric channel and greedy
decoding are shared.

The final models use rotary position embeddings. RoPE applies
position-dependent rotations to queries and keys so attention contains
relative-position information
(\href{https://arxiv.org/abs/2104.09864}{Su et al., 2021}). We selected
it through a matched ablation rather than assuming it would transfer
unchanged to a morphology-factor sequence.

Morphology-flat and the external arms use three global encoder layers.
The signal-preserving composer retains three global encoder layers and
adds its declared composition paths while preserving the matched total
parameter budget of 18.97-million. The historical ablation used a
composer layer plus two global layers.

The shared numeric channel replaces aligned source--target number
occurrences with ordered slots and restores exact strings after
generation. The headline decoder requires every source-present slot
before EOS. This prevents basic number copying from becoming the primary
tokenizer advantage, although forced coverage can extend an otherwise
degenerate output.

\subsection{5.4 Metrics}\label{54-metrics}

We report SacreBLEU, chrF++, exact match and teacher-forced target
cross-entropy. chrF++ includes character and word n-gram overlap
(\href{https://aclanthology.org/W17-4770/}{Popović, 2017}). COMETKiwi is
a reference-free quality estimator based on source and hypothesis rather
than the gold target (\href{https://aclanthology.org/2022.wmt-1.60/}{Rei
et al., 2022}).

We use \textbf{fertility} in its conventional word-normalized sense. If
a corpus contains \(W\) orthographic words, a subword tokenizer with
\(T\) model tokens has token fertility \(T/W\). For morphology-flat, the
corresponding quantity is \textbf{serialized factor fertility}, \(N/W\),
because the model-facing positions are lemmas and grammatical factors
rather than ordinary subword tokens. For the composer,
\textbf{global-state fertility}, \(P/W\), counts the states passed to
sentence-level attention. We report \textbf{positions per grapheme} when
normalizing by Tamil grapheme clusters and \textbf{graphemes per
position} for its inverse. ``Tokens per character'' is avoided because
Unicode code points, UTF-8 bytes and visible Tamil graphemes are not
equivalent.

Preservation measures include number recall/precision, required-slot
recall, a conservative capitalized-reference entity-token proxy, Tamil
leakage and repeated trigrams. Efficiency measures include serialized
and global source positions, updates per second, wall time, peak RSS and
decoding throughput. Paired intervals use 10,000 row-level bootstrap
resamples. They measure test-row uncertainty for one trained seed, not
training-seed variation.

\section{6. Results}\label{6-results}

\subsection{6.1 Development evidence}\label{61-development-evidence}

On the \textbf{69,591-pair development comparison}, composer and flat
were tied:

\begingroup\small\sffamily

{\def\LTcaptype{none} 
\begin{longtable}[]{@{}
  >{\raggedright\arraybackslash}p{(\linewidth - 8\tabcolsep) * \real{0.2800}}
  >{\raggedleft\arraybackslash}p{(\linewidth - 8\tabcolsep) * \real{0.1800}}
  >{\raggedleft\arraybackslash}p{(\linewidth - 8\tabcolsep) * \real{0.1800}}
  >{\raggedleft\arraybackslash}p{(\linewidth - 8\tabcolsep) * \real{0.1800}}
  >{\raggedleft\arraybackslash}p{(\linewidth - 8\tabcolsep) * \real{0.1800}}@{}}
\toprule\noalign{}
\begin{minipage}[b]{\linewidth}\raggedright
Arm
\end{minipage} & \begin{minipage}[b]{\linewidth}\raggedleft
BLEU
\end{minipage} & \begin{minipage}[b]{\linewidth}\raggedleft
chrF++
\end{minipage} & \begin{minipage}[b]{\linewidth}\raggedleft
COMETKiwi
\end{minipage} & \begin{minipage}[b]{\linewidth}\raggedleft
Dev. CE
\end{minipage} \\
\midrule\noalign{}
\endhead
\bottomrule\noalign{}
\endlastfoot
composer & 21.85 & \textbf{48.76} & 0.7050 & 2.4458 \\
morphology-flat & \textbf{22.69} & 48.68 & \textbf{0.7063} &
\textbf{2.4249} \\
AI4Bharat & 21.47 & 47.52 & 0.6951 & 2.5807 \\
Sarvam & 20.80 & 46.51 & 0.6842 & 2.5527 \\
Brahmic & 20.47 & 46.12 & 0.6782 & 2.5343 \\
\end{longtable}
}

\endgroup

Composer-minus-flat sentence-chrF++ was \textbf{+0.185 with a 95\%
interval of {[}-0.262, +0.634{]}}; COMETKiwi was \textbf{-0.00123} with
\textbf{{[}-0.00483, +0.00235{]}}. This justified carrying both systems
to the protected test. Composer reduced mean global semantic states from
\textbf{80.77 to 32.44}. The historical single-summary ablation scored
\textbf{21.65 BLEU, 47.12 chrF++ and 0.6911 COMETKiwi}; its role is
architectural diagnosis rather than a proposed system.

A separate 12,000-update positional ablation compared learned absolute,
scaled fixed sinusoidal and rotary positions. RoPE improved development
loss, BLEU, chrF++, COMETKiwi and repetition behavior for
morphology-flat, the morphology composer and AI4Bharat. We therefore
used one positional system, RoPE, for the final study.

After the protected comparison was complete, we ran development-only
architecture tests to understand composer rather than to revise the
protected result. They produced four main findings:

\begin{itemize}
\tightlist
\item
  models trained without late retrieval were worse for both one- and
  two-summary composers;
\item
  moving retrieval before the final encoder layer was worse than
  retrieving after all encoder layers;
\item
  a hand-written two-summary split preserved information, but the
  decoder made little use of the second state;
\item
  two learned all-factor summaries also became largely redundant and
  required \textbf{37.8\% more global states} than one summary on the
  diagnostic set.
\end{itemize}

\subsection{6.2 Protected quality}\label{62-protected-quality}

\begingroup\small\sffamily

{\def\LTcaptype{none} 
\begin{longtable}[]{@{}
  >{\raggedright\arraybackslash}p{(\linewidth - 8\tabcolsep) * \real{0.2800}}
  >{\raggedleft\arraybackslash}p{(\linewidth - 8\tabcolsep) * \real{0.1800}}
  >{\raggedleft\arraybackslash}p{(\linewidth - 8\tabcolsep) * \real{0.1800}}
  >{\raggedleft\arraybackslash}p{(\linewidth - 8\tabcolsep) * \real{0.1800}}
  >{\raggedleft\arraybackslash}p{(\linewidth - 8\tabcolsep) * \real{0.1800}}@{}}
\toprule\noalign{}
\begin{minipage}[b]{\linewidth}\raggedright
Arm
\end{minipage} & \begin{minipage}[b]{\linewidth}\raggedleft
Loss
\end{minipage} & \begin{minipage}[b]{\linewidth}\raggedleft
BLEU
\end{minipage} & \begin{minipage}[b]{\linewidth}\raggedleft
chrF++
\end{minipage} & \begin{minipage}[b]{\linewidth}\raggedleft
COMETKiwi
\end{minipage} \\
\midrule\noalign{}
\endhead
\bottomrule\noalign{}
\endlastfoot
morphology-flat & 3.8550 & \textbf{10.63} & \textbf{35.26} &
\textbf{0.6276} \\
composer & 3.8519 & 10.30 & 34.88 & 0.6241 \\
AI4Bharat & 4.0137 & 9.92 & 34.16 & 0.6119 \\
Brahmic & 3.8871 & 9.45 & 33.55 & 0.5923 \\
Sarvam & 3.8794 & 9.24 & 33.40 & 0.5985 \\
\end{longtable}
}

\endgroup

Morphology-flat leads all three generation metrics. The omitted
historical ablation had the lowest teacher-forced loss, \textbf{3.8396},
but only \textbf{10.01 BLEU, 34.35 chrF++ and 0.6156 COMETKiwi}. This
illustrates why target cross-entropy is diagnostic rather than the
headline outcome.

Composer-minus-flat has a pooled sentence-chrF++ difference of
\textbf{-0.316 {[}-0.591, -0.037{]}} and COMETKiwi difference
\textbf{-0.00343 {[}-0.00636, -0.00062{]}}. Composer nevertheless
exceeds AI4Bharat by \textbf{+0.743} sentence-chrF++ and
\textbf{+0.01226} COMETKiwi, Sarvam by \textbf{+1.492} and
\textbf{+0.02559}, and Brahmic by \textbf{+1.649 and +0.03184}. All
corresponding paired intervals exclude zero. Against the historical
single-summary ablation, composer gains \textbf{+0.461} sentence-chrF++
and \textbf{+0.00856} COMETKiwi.

The domain breakdown explains the pooled result:

\begingroup\small\sffamily

{\def\LTcaptype{none} 
\begin{longtable}[]{@{}
  >{\raggedright\arraybackslash}p{(\linewidth - 6\tabcolsep) * \real{0.3400}}
  >{\raggedleft\arraybackslash}p{(\linewidth - 6\tabcolsep) * \real{0.2200}}
  >{\raggedleft\arraybackslash}p{(\linewidth - 6\tabcolsep) * \real{0.2200}}
  >{\raggedleft\arraybackslash}p{(\linewidth - 6\tabcolsep) * \real{0.2200}}@{}}
\toprule\noalign{}
\begin{minipage}[b]{\linewidth}\raggedright
Arm
\end{minipage} & \begin{minipage}[b]{\linewidth}\raggedleft
IN22-Gen chrF++ / COMETKiwi
\end{minipage} & \begin{minipage}[b]{\linewidth}\raggedleft
IN22-Conv
\end{minipage} & \begin{minipage}[b]{\linewidth}\raggedleft
FLORES+ devtest
\end{minipage} \\
\midrule\noalign{}
\endhead
\bottomrule\noalign{}
\endlastfoot
morphology-flat & \textbf{36.07 / 0.6229} & \textbf{27.21 / 0.6137} &
\textbf{39.08 / 0.6530} \\
composer-v2 & 35.88 / \textbf{0.6237} & 26.95 / 0.6106 & 38.41 /
0.6446 \\
\end{longtable}
}

\endgroup

The paired intervals include zero on both IN22 partitions. FLORES+
carries the measurable deficit: composer-minus-flat sentence-chrF++ is
-0.754 {[}-1.258, -0.248{]}, and COMETKiwi is -0.00836 {[}-0.01285,
-0.00372{]}.

\subsection{6.3 Preservation and
degeneration}\label{63-preservation-and-degeneration}

\begingroup\small\sffamily

{\def\LTcaptype{none} 
\begin{longtable}[]{@{}
  >{\raggedright\arraybackslash}p{(\linewidth - 8\tabcolsep) * \real{0.2800}}
  >{\raggedleft\arraybackslash}p{(\linewidth - 8\tabcolsep) * \real{0.1800}}
  >{\raggedleft\arraybackslash}p{(\linewidth - 8\tabcolsep) * \real{0.1800}}
  >{\raggedleft\arraybackslash}p{(\linewidth - 8\tabcolsep) * \real{0.1800}}
  >{\raggedleft\arraybackslash}p{(\linewidth - 8\tabcolsep) * \real{0.1800}}@{}}
\toprule\noalign{}
\begin{minipage}[b]{\linewidth}\raggedright
Arm
\end{minipage} & \begin{minipage}[b]{\linewidth}\raggedleft
Number recall
\end{minipage} & \begin{minipage}[b]{\linewidth}\raggedleft
Number precision
\end{minipage} & \begin{minipage}[b]{\linewidth}\raggedleft
Entity proxy
\end{minipage} & \begin{minipage}[b]{\linewidth}\raggedleft
Repeated-trigram rows
\end{minipage} \\
\midrule\noalign{}
\endhead
\bottomrule\noalign{}
\endlastfoot
morphology-flat & \textbf{93.97\%} & 83.60\% & \textbf{39.53\%} &
\textbf{250} \\
composer & \textbf{93.97\%} & \textbf{88.33\%} & 39.38\% & 274 \\
AI4Bharat & 92.90\% & 84.33\% & 37.79\% & 280 \\
Sarvam & 93.26\% & 86.13\% & 35.58\% & 314 \\
Brahmic & 93.62\% & 82.49\% & 35.74\% & 313 \\
\end{longtable}
}

\endgroup

Every arm achieved \textbf{100\% recall of required numeric slots}.
Surface-number scores remain lower because unmatched literals and
additional generated numbers are possible. No arm emitted Tamil script
in its English output. The entity measure is a conservative exact-token
proxy, not full entity evaluation.

\subsection{6.4 Representation length and recorded
runtime}\label{64-representation-length-and-recorded-runtime}

\begingroup\small\sffamily

{\def\LTcaptype{none} 
\begin{longtable}[]{@{}
  >{\raggedright\arraybackslash}p{(\linewidth - 8\tabcolsep) * \real{0.2800}}
  >{\raggedleft\arraybackslash}p{(\linewidth - 8\tabcolsep) * \real{0.1800}}
  >{\raggedleft\arraybackslash}p{(\linewidth - 8\tabcolsep) * \real{0.1800}}
  >{\raggedleft\arraybackslash}p{(\linewidth - 8\tabcolsep) * \real{0.1800}}
  >{\raggedleft\arraybackslash}p{(\linewidth - 8\tabcolsep) * \real{0.1800}}@{}}
\toprule\noalign{}
\begin{minipage}[b]{\linewidth}\raggedright
Arm
\end{minipage} & \begin{minipage}[b]{\linewidth}\raggedleft
Serialized source positions/row
\end{minipage} & \begin{minipage}[b]{\linewidth}\raggedleft
Global encoder states/row
\end{minipage} & \begin{minipage}[b]{\linewidth}\raggedleft
Recorded rows/s
\end{minipage} & \begin{minipage}[b]{\linewidth}\raggedleft
Peak RSS
\end{minipage} \\
\midrule\noalign{}
\endhead
\bottomrule\noalign{}
\endlastfoot
morphology-flat & 71.48 & 71.48 & 4.12 & 6,544 MiB \\
composer & 71.48 & \textbf{29.08} & 3.15 & 7,254 MiB \\
AI4Bharat & \textbf{25.45} & 25.45 & 4.50 & 6,754 MiB \\
Sarvam & 32.88 & 32.88 & 3.27 & \textbf{6,135 MiB} \\
Brahmic & 45.52 & 45.52 & \textbf{6.14} & 6,228 MiB \\
\end{longtable}
}

\endgroup

Composer reduces \textbf{252,979 protected raw factors to 102,897 global
states} (by \textbf{59.3\%}). Our training and inference runs were not
repeated under controlled system load, so they do not establish
comparable wall-clock speeds. The operation-count estimate in Section
4.2 instead predicts fewer FLOPs for composer.

\subsection{6.5 Evaluation-only vocabulary
IDs}\label{65-evaluation-only-vocabulary-ids}

No tokenizer emitted its unknown ID. Each frozen dense map nevertheless
encountered native IDs absent from training:

\begingroup\small\sffamily

{\def\LTcaptype{none} 
\begin{longtable}[]{@{}
  >{\raggedright\arraybackslash}p{(\linewidth - 6\tabcolsep) * \real{0.3400}}
  >{\raggedleft\arraybackslash}p{(\linewidth - 6\tabcolsep) * \real{0.2200}}
  >{\raggedleft\arraybackslash}p{(\linewidth - 6\tabcolsep) * \real{0.2200}}
  >{\raggedleft\arraybackslash}p{(\linewidth - 6\tabcolsep) * \real{0.2200}}@{}}
\toprule\noalign{}
\begin{minipage}[b]{\linewidth}\raggedright
Arm
\end{minipage} & \begin{minipage}[b]{\linewidth}\raggedleft
Distinct IDs
\end{minipage} & \begin{minipage}[b]{\linewidth}\raggedleft
Occurrences
\end{minipage} & \begin{minipage}[b]{\linewidth}\raggedleft
Rows affected
\end{minipage} \\
\midrule\noalign{}
\endhead
\bottomrule\noalign{}
\endlastfoot
morphology & 298 & 423 & 344 (9.72\%) \\
AI4Bharat & 366 & 627 & 527 (14.89\%) \\
Sarvam & 63 & 230 & 164 (4.63\%) \\
Brahmic & 82 & 112 & 78 (2.20\%) \\
\end{longtable}
}

\endgroup

These IDs use initialized but untrained embedding rows. This is
preferable to collapsing the surface to
\texttt{\textless{}unk\textgreater{}}, but it remains a generalization
weakness. Future systems should initialize unseen lemma or grapheme rows
from spelling, features or lexical priors.

\subsection{6.6 Qualitative pattern}\label{66-qualitative-pattern}

The frozen AI review found that composer often handles short
conversational constructions and common paraphrases well. Flat more
often preserves a specific lexical item, proper name, organization or
long formal clause. Composer can turn compressed factors into a
plausible but incorrect lexical choice. Difficult long inputs expose
omissions and occasional repetition in both small systems. The
composer's one grammar summary can still be a bottleneck despite its
direct lexical path. However, the tested two-summary replacements did
not solve that problem: they added weakly used or redundant states.

\section{7. Discussion}\label{7-discussion}

\subsection{7.1 What the comparison
supports}\label{71-what-the-comparison-supports}

Morphology-flat versus each external flat arm is the cleanest tokenizer
contrast. Under the shared model and budget, the complete morphology
representation provides useful translation signal. The experiment does
not attribute the gain to case, tense, lemma reuse or any single
feature. A future causal ablation should compare full morphology,
lemma-only, lemma-plus-POS, shuffled feature labels and a surface-only
word composer.

Composer versus morphology-flat is an architecture contrast. The main
result shows that direct lexical, grammar and fallback paths preserve
most flat-model quality while substantially shortening global attention.
The historical single-summary ablation confirms that efficiency can be
purchased by discarding information.

The later architecture tests clarify what did and did not repair the
compact model. Late retrieval is useful, and placing it after all
encoder layers works better than placing another encoder layer after it.
Simply adding a second summary is not enough. The hand-written split
produced a sparse secondary channel, while two learned all-factor
summaries converged toward similar states. A future design would need a
clearer source of complementary information---such as a conditional
residual path for unusually complex words---and would still need to beat
the one-summary composer control under matched quality and runtime
tests.

\subsection{7.2 Efficiency is
multidimensional}\label{72-efficiency-is-multidimensional}

Serialized factor fertility, global-state fertility, training
throughput, memory and decoding speed are related but not
interchangeable. Flat morphology has the best quality but the highest
factor fertility---that is, the most serialized model positions per
source word---and the longest global sequence. Composer substantially
lowers global-state fertility and requires less estimated arithmetic.
AI4Bharat uses the shortest mean source sequence, while Brahmic had the
highest recorded protected row throughput. A claim of compute
superiority would require repeated controlled benchmarks and matched
token, wall-time or FLOP budgets rather than matched examples and
updates alone.

\subsection{7.3 Coverage is not one
number}\label{73-coverage-is-not-one-number}

Exact reversibility, direct FST coverage, entity coverage, grapheme
fallback, byte fallback, unknown IDs and learned embeddings are
different properties. The protected evaluation demonstrates the last
distinction particularly clearly: a tokenizer can preserve a valid
unseen ID without giving the model a trained meaning for it.

The morphology resource should therefore continue to publish multiple
measurements rather than a synthetic ``tokenizer score.'' Classical
Tamil, names, code mixing, rare words and noisy text remain important
coverage areas. Context-sensitive analysis selection also remains open.

\subsection{7.4 Domain knowledge and the Bitter
Lesson}\label{74-domain-knowledge-and-the-bitter-lesson}

Sutton\textquotesingle s
\href{http://www.incompleteideas.net/IncIdeas/BitterLesson.html}{Bitter
Lesson} argues that, over long periods, general methods that can make
use of increasing computation tend to outperform systems built around
human knowledge of a particular domain. Our approach is deliberately in
tension with that lesson. The FSTs, lexical classes, semantic factors
and word-local attention structure all place Tamil-specific knowledge in
front of the learned model.

Our results do not overturn the Bitter Lesson. They show that this
knowledge helped under one deliberately small and fixed budget:
\textbf{69,591 training pairs}, an \textbf{18.97-million-parameter
model} and \textbf{40,000 updates}. We did not measure a scaling curve.
A much larger general model with much more Tamil data might learn the
same regularities from use, and the advantage of the explicit
representation might shrink or reverse. The external arms also test
tokenizer interfaces inside our small model, not the complete pretrained
systems from which those tokenizers came.

There are nevertheless reasons to test linguistic structure rather than
assume that scale will always be available. Tamil is relatively
data-limited; the flat result suggests that explicit factors can improve
sample efficiency in that setting. Exact reconstruction, readable
analyses, controlled fallback and auditable grammar are also system
properties, not merely shortcuts to a benchmark score. The word-composer
asks a more scale-compatible question: whether those properties can be
retained while reducing the computation spent on the expanded factor
stream.

The decisive experiment is therefore not domain knowledge versus scale
in the abstract. It is a scaling study across several data, parameter
and FLOP budgets. Such a study should test whether
morphology\textquotesingle s quality advantage persists, narrows or
reverses, and whether the composer converts any remaining advantage into
lower total computation. Until then, the appropriate claim is limited:
Tamil-specific structure is useful in the low-resource, small-model
regime measured here, but it has not been shown to dominate general
methods at scale.

\section{8. Limitations}\label{8-limitations}

The study has one primary language, one translation direction, small
randomly initialized models and one training seed. Bootstrap intervals
measure variation over rows, not seed stability. The final test was
protected from model selection, but it does not replace multi-seed
replication.

The later ablations were run after the six-arm protected comparison and
used development data only. They help explain the architecture, but they
do not have new protected-test scores and should not be presented as
protected comparisons.

The training union includes synthetic Tamil. Automatic filtering and AI
review reduce obvious failures but do not replace independent bilingual
evaluation. COMETKiwi is reference-free and imperfect. The entity metric
is only a proxy. The required-number decoder changes generation behavior
and can extend weak outputs.

The external arms test selected tokenizer artifacts inside our
architecture, not complete Sarvam, AI4Bharat or Brahmic systems. Matched
parameters, examples and updates do not match processed source tokens,
FLOPs or wall time. Local CPU behavior may not predict optimized GPU
kernels.

Finally, the FSTs are broad but incomplete. They have poor coverage on
Classical Tamil. Deterministic context-free ranking can select the wrong
valid reading. The exact public codec is designed for auditability and
reversibility, not asserted to be the most compact possible model
interface.

\section{9. Ethics, provenance and
licensing}\label{9-ethics-provenance-and-licensing}

The release must retain attribution to the upstream ThamizhiMorph models
and their authors, Kengatharaiyer Sarveswaran, Gihan Dias and Miriam
Butt, as well as every lexical, corpus and model source. Component
licenses differ: Apache-2.0 code and tokenizer artifacts must not be
conflated with CC BY-SA lexical or documentary material. Private or
license-restricted word lists and audit inventories are excluded from
public archives. Dataset text is released only where the corresponding
license permits it; otherwise the project publishes identifiers, hashes
and reconstruction scripts.

Synthetic sources are labeled as synthetic and retain their teacher
model, revision and generation provenance. They should not be presented
as native human-authored Tamil.

OpenAI Codex, primarily using the GPT-5.6 Sol model, served as an AI
research and coding assistant. It helped implement and test software,
audit data, monitor experiments, prepare figures and tables, inspect
qualitative outputs and review prose. This work is credited as AI
assistance, not as bilingual human annotation or paper authorship. Anand
Murugan selected the research questions, approved the experimental and
release decisions, reviewed and edited the article, and accepts
responsibility for the published claims.

All reported training and evaluation ran locally with zero incremental
paid external compute cost. Hardware and approximate wall time should
accompany the release so that the resource cost remains visible. The
small research translation checkpoints are not suitable for
authoritative deployment. Omissions, altered relations and plausible
hallucinations remain common enough to create harm if outputs are
treated as reliable translations.

\section{10. Artifacts, provenance and
release}\label{10-artifacts-provenance-and-release}

The project is divided by responsibility:

\begingroup\small\sffamily

{\def\LTcaptype{none} 
\begin{longtable}[]{@{}
  >{\raggedright\arraybackslash}p{(\linewidth - 2\tabcolsep) * \real{0.3200}}
  >{\raggedright\arraybackslash}p{(\linewidth - 2\tabcolsep) * \real{0.6800}}@{}}
\toprule\noalign{}
\begin{minipage}[b]{\linewidth}\raggedright
Resource
\end{minipage} & \begin{minipage}[b]{\linewidth}\raggedright
Public location
\end{minipage} \\
\midrule\noalign{}
\endhead
\bottomrule\noalign{}
\endlastfoot
original ThamizhiMorph FST models &
\href{https://github.com/sarves/thamizhi-morph}{Open} \\
tokenizer, codec, runtime FSTs, tests &
\href{https://github.com/Indic-AI-Experiments/tamil-morph-tokenizer}{Open} \\
versioned morphology/FST release &
\href{https://github.com/Indic-AI-Experiments/tamil-morphology}{Open} \\
experiment code, configs and analyses &
\href{https://github.com/Indic-AI-Experiments/tamil-tokenizer-experiments}{Open} \\
patched build lineage and application &
\href{https://github.com/kupilikula/Solladukku}{Open} \\
interactive tokenizer web interface &
\href{https://anandmurugan.me/resources/tamil-tokenizer}{Open} \\
searchable grammatical and semantic label vocabulary &
\href{https://anandmurugan.me/resources/tamil-tokenizer/vocabulary}{Open} \\
machine-readable grammatical and semantic label vocabulary &
\href{https://github.com/Indic-AI-Experiments/tamil-morph-tokenizer/blob/main/tamil_morph_tokenizer/data/vocabulary/semantic_token_reference.json}{Open} \\
\end{longtable}
}

\endgroup

Tokenizer artifacts are frozen at
\texttt{\seqsplit{789570741ed50321753911e6d7233dd2114bcece}}. The
manifest calls this \texttt{0.1.0-rc10}; stale Python metadata calls it
\texttt{0.1.0rc8}. Release \texttt{0.1.0-rc11} corrects metadata only;
vocabulary and IDs are unchanged. Morphology is frozen at
\texttt{\seqsplit{c9fe57cca09b0c6f51178386261232bc699dcc9a}}
(\texttt{0.1.0-rc8}). External tokenizer revisions are:

\begingroup\small\sffamily

{\def\LTcaptype{none} 
\begin{longtable}[]{@{}
  >{\raggedright\arraybackslash}p{(\linewidth - 2\tabcolsep) * \real{0.3000}}
  >{\raggedright\arraybackslash}p{(\linewidth - 2\tabcolsep) * \real{0.7000}}@{}}
\toprule\noalign{}
\begin{minipage}[b]{\linewidth}\raggedright
Artifact
\end{minipage} & \begin{minipage}[b]{\linewidth}\raggedright
Frozen revision
\end{minipage} \\
\midrule\noalign{}
\endhead
\bottomrule\noalign{}
\endlastfoot
\texttt{sarvamai/sarvam-1} &
\texttt{\seqsplit{e9607337286ddf496d4a2562b194e489dcf3feea}} \\
\texttt{ai4bharat/IndicBERTv2-MLM-Sam-TLM} &
\texttt{\seqsplit{bb783337859e3d7957de5ba82766ddea51e8fc3e}} \\
\texttt{\seqsplit{theschoolofai/BrahmicTokenizer-131K}} &
\texttt{\seqsplit{93df154cbc9dbf038a222c010d9b43906a8a72c3}} \\
\end{longtable}
}

\endgroup

Protected dataset revisions are
\texttt{\seqsplit{e042ab3d3063110b1a85efa0a59bdbf8553bb928}} for
IN22-Gen, \texttt{\seqsplit{18cd45870ff0a9e65df9b80dbbcc615eec0e4899}}
for IN22-Conv and
\texttt{\seqsplit{5fec6c13f9e5a4db2f745d4ec0d7c9721ddc4f0}} for FLORES+.
File-level hashes remain in the machine-readable manifest rather than
being duplicated in the main text.

The public repositories contain machine-readable manifests that identify
the released source commits and record checksums for the FSTs,
vocabulary and token mappings. The experiment manifest also records the
configuration, software environment, hardware, commands and cost of the
comparison, together with checksums for the \textbf{69,591-pair training
set} and all \textbf{six trained checkpoints}. The published material
supports three levels of reproduction: testing the tokenizer and its
exact reconstruction, recomputing the reported analysis from
redistributable artifacts and aggregate evidence, or repeating the
complete training procedure with the released code and configuration.

The protected IN22 and FLORES+ source sentences, reference translations
and row-level predictions are not redistributed in this release. This
respects the datasets\textquotesingle{} individual license conditions
and preserves the evaluation boundary used in the study. The release
instead records the exact dataset revisions and checksums, provides the
permitted preparation and evaluation code, and publishes the aggregate
results reported in this paper.

\section{11. Conclusion}\label{11-conclusion}

Tamil morphology can be exposed through a deterministic, auditable and
byte-exact tokenizer. In the controlled experiment, the flat morphology
representation produces the best protected Tamil--English translation
quality, outperforming the selected external-tokenizer arms at the cost
of higher serialized factor fertility: more model positions per source
word and, consequently, a longer mean source sequence.

The hierarchical results refine that finding. The signal-preserving
composer keeps lexical identity, grammar and fallback on separate paths,
exceeds all external arms and reduces global source states by
\textbf{59.3\%}. It does not fully match flat on long formal sentences.
The FLOP estimates predict less arithmetic though the wall-time
comparison was not controlled. A historical aggressive-compression
ablation confirms that removing direct lexical information loses
quality. Matched controls show that late retrieval is useful. Earlier
retrieval and both tested two-summary designs fail to improve the
overall tradeoff, making our presented composer the best supported
compact model rather than merely the first one tried.

The resulting claim is not that linguistic tokenization always wins. It
is that explicit Tamil morphology provides useful reusable signal, and
that tokenization and attention topology should be designed together. A
hierarchy is valuable only when it preserves the distinctions it was
introduced to organize.

\clearpage\begin{landscape}

\section{Appendix A. Complete grammatical and semantic label
vocabulary}\label{appendix-a-complete-grammatical-and-semantic-label-vocabulary}

The released vocabulary contains \textbf{222 fixed grammatical and
semantic labels}. They cover grammar, relations and named-entity types.
Ordinary lemmas are a separate part of the unified model vocabulary.
This includes \textbf{58 words} used as secondary lemmas in multi-lemma
and auxiliary analyses; they remain lexical states rather than becoming
a special semantic-token class.

Token type is recorded explicitly rather than inferred from numerical
position. For compatibility with the trained checkpoints, secondary
lemmas retain IDs 969--1026. The nearby IDs 965--968 are \textbf{4
ordinary noun lemmas} left there by an older incremental refresh. None
of these lexical entries is included in the \textbf{222-label} count or
in the table below.

The table also excludes codec structure, spelling-reconstruction
markers, grapheme or byte fallback, and the ordinary lemma vocabulary.
The token IDs below are those of the frozen \texttt{0.1.0-rc10} stream
and remain unchanged in \texttt{0.1.0-rc11}. A searchable copy is also
published with the
\href{https://anandmurugan.me/resources/tamil-tokenizer/vocabulary}{interactive
tokenizer}.

\begingroup\scriptsize\sffamily

{\def\LTcaptype{none} 
\begin{longtable}[]{@{}
  >{\raggedleft\arraybackslash}p{(\linewidth - 6\tabcolsep) * \real{0.0600}}
  >{\raggedright\arraybackslash}p{(\linewidth - 6\tabcolsep) * \real{0.2400}}
  >{\raggedright\arraybackslash}p{(\linewidth - 6\tabcolsep) * \real{0.2200}}
  >{\raggedright\arraybackslash}p{(\linewidth - 6\tabcolsep) * \real{0.4800}}@{}}
\toprule\noalign{}
\begin{minipage}[b]{\linewidth}\raggedleft
ID
\end{minipage} & \begin{minipage}[b]{\linewidth}\raggedright
Token
\end{minipage} & \begin{minipage}[b]{\linewidth}\raggedright
Family
\end{minipage} & \begin{minipage}[b]{\linewidth}\raggedright
Meaning
\end{minipage} \\
\midrule\noalign{}
\endhead
\bottomrule\noalign{}
\endlastfoot
743 & \texttt{\textless{}ABBREVIATION\textgreater{}} & grammatical or
semantic feature & Marks an abbreviation. \\
744 & \texttt{\textless{}ACTION\_NOMINAL\textgreater{}} & grammatical or
semantic feature & Marks a verb-derived noun that names an action or
event. \\
745 & \texttt{\textless{}ADJECTIVAL\_PARTICIPLE\textgreater{}} &
participle & Marks a verb form used to modify a noun. \\
746 & \texttt{\textless{}ASPECT\_PERFECT\textgreater{}} & aspect & Marks
a completed action or a resulting state. \\
747 & \texttt{\textless{}ASPECT\_PROSPECTIVE\textgreater{}} & aspect &
Marks an action viewed as expected or about to happen. \\
748 & \texttt{\textless{}AUX\_ATTITUDINAL\textgreater{}} & grammatical
or semantic feature & Broad inherited FST label for an auxiliary
construction that expresses the speaker\textquotesingle s stance. \\
749 & \texttt{\textless{}CASE\_ABL\textgreater{}} & case & Ablative
case: from, out of, or away from. \\
750 & \texttt{\textless{}CASE\_ACC\textgreater{}} & case & Accusative
case: usually the direct object. \\
751 & \texttt{\textless{}CASE\_DAT\textgreater{}} & case & Dative case:
usually to or for. \\
752 & \texttt{\textless{}CASE\_GEN\textgreater{}} & case & Genitive
case: possession or an of-relation. \\
753 & \texttt{\textless{}CASE\_INST\textgreater{}} & case & Instrumental
case: by, with, or using. \\
754 & \texttt{\textless{}CASE\_LOC\textgreater{}} & case & Locative
case: in, at, or on. \\
755 & \texttt{\textless{}CASE\_MARKER\textgreater{}} & case & Broad
inherited FST label indicating that a case marker is present. \\
756 & \texttt{\textless{}CASE\_NOM\textgreater{}} & case & Nominative or
unmarked base case, often used for the subject. \\
757 & \texttt{\textless{}CASE\_SOC\textgreater{}} & case & Sociative
case: with or together with. \\
758 & \texttt{\textless{}CASE\_TRANS\textgreater{}} & case & Translative
or adverbial case-like form: as, becoming, or in a stated manner. \\
759 & \texttt{\textless{}CASE\_VOC\textgreater{}} & case & Vocative case
used for direct address. \\
760 & \texttt{\textless{}CLITIC\_ADD\textgreater{}} & clitic & Additive
clitic: also, too, or and. \\
761 & \texttt{\textless{}CLITIC\_FOCUS\textgreater{}} & clitic & Focus
or emphatic clitic, often corresponding to தான். \\
762 & \texttt{\textless{}COMPARATIVE\textgreater{}} & grammatical or
semantic feature & Marks a comparison such as than, more, or less. \\
763 & \texttt{\textless{}COMPLEMENTIZER\textgreater{}} & grammatical or
semantic feature & Introduces a quoted, reported, or embedded clause. \\
764 & \texttt{\textless{}COMPOUND\_MODIFIER\textgreater{}} & grammatical
or semantic feature & Marks a noun used attributively before another
word in a compound. \\
765 & \texttt{\textless{}COPULA\textgreater{}} & grammatical or semantic
feature & Marks a copular expression that links a subject with a
description or identity. \\
766 & \texttt{\textless{}COP\_BECOME\textgreater{}} & grammatical or
semantic feature & Marks a change into a state: become. \\
767 & \texttt{\textless{}DEGREE\textgreater{}} & grammatical or semantic
feature & Marks an amount or degree expression. \\
768 & \texttt{\textless{}DEICTIC\textgreater{}} & deixis & General
demonstrative or pointing meaning. \\
769 & \texttt{\textless{}DEICTIC\_DIST\textgreater{}} & deixis & Distal
demonstrative: that, there, or then. \\
770 & \texttt{\textless{}DEICTIC\_INTERROGATIVE\textgreater{}} & deixis
& Interrogative demonstrative: which, where, or when. \\
771 & \texttt{\textless{}DEICTIC\_MED\textgreater{}} & deixis & Medial
demonstrative: an intermediate distance. \\
772 & \texttt{\textless{}DEICTIC\_PROX\textgreater{}} & deixis &
Proximal demonstrative: this, here, or now. \\
773 & \texttt{\textless{}DEICTIC\_SAME\textgreater{}} & deixis & Marks
identity or sameness: the same. \\
774 & \texttt{\textless{}DEICTIC\_SITUATION\textgreater{}} & deixis &
Points to a situation or context. \\
775 & \texttt{\textless{}DEICTIC\_TIME\textgreater{}} & deixis & Points
to a time. \\
776 & \texttt{\textless{}DEICTIC\_TYPE\textgreater{}} & deixis & Points
to a kind or type. \\
777 & \texttt{\textless{}DERIV\_AATTAM\textgreater{}} & grammatical or
semantic feature & Marks the ஆட்டம்-derived manner or likeness
construction. \\
778 & \texttt{\textless{}DETERMINER\textgreater{}} & grammatical or
semantic feature & Marks a word that specifies or limits a noun. \\
779 & \texttt{\textless{}DISTRIBUTIVE\textgreater{}} & grammatical or
semantic feature & Distributive meaning: each, respective, or one by
one. \\
780 & \texttt{\textless{}ENTITY\_BRAND\textgreater{}} & named entity &
Named entity: brand. \\
781 & \texttt{\textless{}ENTITY\_CITY\textgreater{}} & named entity &
Named entity: city. \\
782 & \texttt{\textless{}ENTITY\_COUNTRY\textgreater{}} & named entity &
Named entity: country. \\
783 & \texttt{\textless{}ENTITY\_ORG\textgreater{}} & named entity &
Named entity: organization. \\
784 & \texttt{\textless{}ENTITY\_OTHER\textgreater{}} & named entity &
Named entity: other reviewed entity type. \\
785 & \texttt{\textless{}ENTITY\_PERSON\textgreater{}} & named entity &
Named entity: person. \\
786 & \texttt{\textless{}ENTITY\_PLACE\textgreater{}} & named entity &
Named entity: place. \\
787 & \texttt{\textless{}ENTITY\_REGION\textgreater{}} & named entity &
Named entity: region. \\
788 & \texttt{\textless{}ENTITY\_WORK\textgreater{}} & named entity &
Named entity: named creative work. \\
789 & \texttt{\textless{}EUPHONIC\_AUGMENT\textgreater{}} & grammatical
or semantic feature & Marks an inserted sound used to join morphemes
smoothly. \\
790 & \texttt{\textless{}EVIDENTIAL\_REPORTATIVE\textgreater{}} &
grammatical or semantic feature & Marks information presented as
reported rather than directly witnessed. \\
791 & \texttt{\textless{}EXISTENTIAL\textgreater{}} & grammatical or
semantic feature & Marks existence or availability. \\
792 & \texttt{\textless{}FUTURE\_ADJECTIVAL\_PARTICIPLE\textgreater{}} &
participle & Marks a future-oriented verb form used to modify a noun. \\
793 & \texttt{\textless{}INDEFINITE\textgreater{}} & grammatical or
semantic feature & Marks an indefinite meaning such as some or any. \\
794 & \texttt{\textless{}LETTER\_NAME\textgreater{}} & grammatical or
semantic feature & Marks a spoken or written letter name. \\
795 & \texttt{\textless{}MANNER\_PURPOSE\textgreater{}} & grammatical or
semantic feature & Marks a directed manner or intended outcome, often in
-உமாறு. \\
796 & \texttt{\textless{}MEASUREMENT\_UNIT\textgreater{}} & grammatical
or semantic feature & Marks a unit of measurement. \\
797 & \texttt{\textless{}MODAL\textgreater{}} & modality & Broad modal
meaning such as ability, necessity, or possibility. \\
798 & \texttt{\textless{}MODAL\_MUST\textgreater{}} & modality &
Necessity or obligation: must, should, or need to. \\
799 & \texttt{\textless{}MODAL\_WORTHY\textgreater{}} & modality & Marks
suitability or worthiness. \\
800 & \texttt{\textless{}MOOD\_CONDITIONAL\textgreater{}} & mood &
Conditional mood: if or under a condition. \\
801 & \texttt{\textless{}MOOD\_OPTATIVE\textgreater{}} & mood & Optative
mood: a wish, hope, or blessing. \\
802 & \texttt{\textless{}MOOD\_PARTICLE\textgreater{}} & mood & Marks a
particle that contributes mood. \\
803 & \texttt{\textless{}MOOD\_PROHIBITIVE\textgreater{}} & mood &
Negative command: do not. \\
804 & \texttt{\textless{}MOOD\_QUESTION\textgreater{}} & mood & Marks a
question. \\
805 & \texttt{\textless{}MORPH\_AFFIRM\textgreater{}} & legacy FST label
& Legacy FST label for affirmative meaning; retained as a fixed readable
factor rather than created dynamically. \\
806 & \texttt{\textless{}MORPH\_ALT\textgreater{}} & legacy FST label &
Legacy FST label for alternative form or reading; retained as a fixed
readable factor rather than created dynamically. \\
807 & \texttt{\textless{}MORPH\_BEN\textgreater{}} & legacy FST label &
Legacy FST label for benefactive meaning; retained as a fixed readable
factor rather than created dynamically. \\
808 & \texttt{\textless{}MORPH\_CMPR\textgreater{}} & legacy FST label &
Legacy FST label for comparative meaning; retained as a fixed readable
factor rather than created dynamically. \\
809 & \texttt{\textless{}MORPH\_CONJUNCTION\textgreater{}} & legacy FST
label & Legacy FST label for conjunction; retained as a fixed readable
factor rather than created dynamically. \\
810 & \texttt{\textless{}MORPH\_DEICTIC\textgreater{}} & legacy FST
label & Legacy FST label for deictic or demonstrative meaning; retained
as a fixed readable factor rather than created dynamically. \\
811 & \texttt{\textless{}MORPH\_EXCLAM\textgreater{}} & legacy FST label
& Legacy FST label for exclamation; retained as a fixed readable factor
rather than created dynamically. \\
812 & \texttt{\textless{}MORPH\_INT\textgreater{}} & legacy FST label &
Legacy FST label for intensifying or interrogative legacy label;
retained as a fixed readable factor rather than created dynamically. \\
813 & \texttt{\textless{}MORPH\_INTERJECTION\textgreater{}} & legacy FST
label & Legacy FST label for interjection; retained as a fixed readable
factor rather than created dynamically. \\
814 & \texttt{\textless{}MORPH\_INTERROGATIVE\textgreater{}} & legacy
FST label & Legacy FST label for interrogative meaning; retained as a
fixed readable factor rather than created dynamically. \\
815 & \texttt{\textless{}MORPH\_LIMIT\textgreater{}} & legacy FST label
& Legacy FST label for limit or restriction; retained as a fixed
readable factor rather than created dynamically. \\
816 & \texttt{\textless{}MORPH\_LOAN\textgreater{}} & legacy FST label &
Legacy FST label for loanword; retained as a fixed readable factor
rather than created dynamically. \\
817 & \texttt{\textless{}MORPH\_NEUT\textgreater{}} & legacy FST label &
Legacy FST label for neuter agreement or class; retained as a fixed
readable factor rather than created dynamically. \\
818 & \texttt{\textless{}MORPH\_N\_PATHIL\textgreater{}} & legacy FST
label & Legacy FST label for noun-based பதில் relational construction;
retained as a fixed readable factor rather than created dynamically. \\
819 & \texttt{\textless{}MORPH\_OTHER\textgreater{}} & legacy FST label
& Legacy FST label for other inherited FST category; retained as a fixed
readable factor rather than created dynamically. \\
820 & \texttt{\textless{}MORPH\_PRIV\textgreater{}} & legacy FST label &
Legacy FST label for privative meaning; retained as a fixed readable
factor rather than created dynamically. \\
821 & \texttt{\textless{}MORPH\_PSP\_ALLAAMAL\textgreater{}} & legacy
postposition & Legacy FST postposition label meaning without. It remains
fixed in the released vocabulary pending a narrower named mapping. \\
822 & \texttt{\textless{}MORPH\_PSP\_APPAAL\textgreater{}} & legacy
postposition & Legacy FST postposition label meaning beyond or on the
other side. It remains fixed in the released vocabulary pending a
narrower named mapping. \\
823 & \texttt{\textless{}MORPH\_PSP\_APPAAL\_IRUNTU\textgreater{}} &
legacy postposition & Legacy FST postposition label meaning from beyond
or on the other side. It remains fixed in the released vocabulary
pending a narrower named mapping. \\
824 & \texttt{\textless{}MORPH\_PSP\_APPURAM\textgreater{}} & legacy
postposition & Legacy FST postposition label meaning after. It remains
fixed in the released vocabulary pending a narrower named mapping. \\
825 & \texttt{\textless{}MORPH\_PSP\_ARUKIL\textgreater{}} & legacy
postposition & Legacy FST postposition label meaning near. It remains
fixed in the released vocabulary pending a narrower named mapping. \\
826 & \texttt{\textless{}MORPH\_PSP\_ARUKIL\_IRUNTU\textgreater{}} &
legacy postposition & Legacy FST postposition label meaning from near.
It remains fixed in the released vocabulary pending a narrower named
mapping. \\
827 & \texttt{\textless{}MORPH\_PSP\_ATIYIL\textgreater{}} & legacy
postposition & Legacy FST postposition label meaning under or at the
foot of. It remains fixed in the released vocabulary pending a narrower
named mapping. \\
828 & \texttt{\textless{}MORPH\_PSP\_ATIYIL\_IRUNTU\textgreater{}} &
legacy postposition & Legacy FST postposition label meaning from under
or at the foot of. It remains fixed in the released vocabulary pending a
narrower named mapping. \\
829 & \texttt{\textless{}MORPH\_PSP\_ETHIR\textgreater{}} & legacy
postposition & Legacy FST postposition label meaning opposite or
against. It remains fixed in the released vocabulary pending a narrower
named mapping. \\
830 & \texttt{\textless{}MORPH\_PSP\_ETHIRE\textgreater{}} & legacy
postposition & Legacy FST postposition label meaning opposite or facing.
It remains fixed in the released vocabulary pending a narrower named
mapping. \\
831 & \texttt{\textless{}MORPH\_PSP\_ETHIRE\_IRUNTU\textgreater{}} &
legacy postposition & Legacy FST postposition label meaning from
opposite or facing. It remains fixed in the released vocabulary pending
a narrower named mapping. \\
832 & \texttt{\textless{}MORPH\_PSP\_ETHIR\_IRUNTU\textgreater{}} &
legacy postposition & Legacy FST postposition label meaning from
opposite or against. It remains fixed in the released vocabulary pending
a narrower named mapping. \\
833 & \texttt{\textless{}MORPH\_PSP\_IDAIYIL\textgreater{}} & legacy
postposition & Legacy FST postposition label meaning between or among.
It remains fixed in the released vocabulary pending a narrower named
mapping. \\
834 & \texttt{\textless{}MORPH\_PSP\_IDAIYL\textgreater{}} & legacy
postposition & Legacy FST postposition label meaning between or among.
It remains fixed in the released vocabulary pending a narrower named
mapping. \\
835 & \texttt{\textless{}MORPH\_PSP\_IDAYIL\_IRUNTU\textgreater{}} &
legacy postposition & Legacy FST postposition label meaning from between
or among. It remains fixed in the released vocabulary pending a narrower
named mapping. \\
836 & \texttt{\textless{}MORPH\_PSP\_ILLAAMAL\textgreater{}} & legacy
postposition & Legacy FST postposition label meaning without. It remains
fixed in the released vocabulary pending a narrower named mapping. \\
837 & \texttt{\textless{}MORPH\_PSP\_KEEL\textgreater{}} & legacy
postposition & Legacy FST postposition label meaning below. It remains
fixed in the released vocabulary pending a narrower named mapping. \\
838 & \texttt{\textless{}MORPH\_PSP\_KEELE\textgreater{}} & legacy
postposition & Legacy FST postposition label meaning below. It remains
fixed in the released vocabulary pending a narrower named mapping. \\
839 & \texttt{\textless{}MORPH\_PSP\_KEELE\_IRUNTU\textgreater{}} &
legacy postposition & Legacy FST postposition label meaning from below.
It remains fixed in the released vocabulary pending a narrower named
mapping. \\
840 & \texttt{\textless{}MORPH\_PSP\_KEEL\_IRUNTU\textgreater{}} &
legacy postposition & Legacy FST postposition label meaning from below.
It remains fixed in the released vocabulary pending a narrower named
mapping. \\
841 & \texttt{\textless{}MORPH\_PSP\_KURUKKE\textgreater{}} & legacy
postposition & Legacy FST postposition label meaning across. It remains
fixed in the released vocabulary pending a narrower named mapping. \\
842 & \texttt{\textless{}MORPH\_PSP\_KURUKKE\_IRUNTU\textgreater{}} &
legacy postposition & Legacy FST postposition label meaning from across.
It remains fixed in the released vocabulary pending a narrower named
mapping. \\
843 & \texttt{\textless{}MORPH\_PSP\_MEEL\textgreater{}} & legacy
postposition & Legacy FST postposition label meaning above or on. It
remains fixed in the released vocabulary pending a narrower named
mapping. \\
844 & \texttt{\textless{}MORPH\_PSP\_MEELE\textgreater{}} & legacy
postposition & Legacy FST postposition label meaning above or on. It
remains fixed in the released vocabulary pending a narrower named
mapping. \\
845 & \texttt{\textless{}MORPH\_PSP\_MEELE\_IRUNTU\textgreater{}} &
legacy postposition & Legacy FST postposition label meaning from above
or on. It remains fixed in the released vocabulary pending a narrower
named mapping. \\
846 & \texttt{\textless{}MORPH\_PSP\_MEEL\_IRUNTU\textgreater{}} &
legacy postposition & Legacy FST postposition label meaning from above
or on. It remains fixed in the released vocabulary pending a narrower
named mapping. \\
847 & \texttt{\textless{}MORPH\_PSP\_MUN\textgreater{}} & legacy
postposition & Legacy FST postposition label meaning before or in front
of. It remains fixed in the released vocabulary pending a narrower named
mapping. \\
848 & \texttt{\textless{}MORPH\_PSP\_MUNNAAL\textgreater{}} & legacy
postposition & Legacy FST postposition label meaning before. It remains
fixed in the released vocabulary pending a narrower named mapping. \\
849 & \texttt{\textless{}MORPH\_PSP\_MUNNAAL\_IRUNTU\textgreater{}} &
legacy postposition & Legacy FST postposition label meaning from before.
It remains fixed in the released vocabulary pending a narrower named
mapping. \\
850 & \texttt{\textless{}MORPH\_PSP\_MUNNE\textgreater{}} & legacy
postposition & Legacy FST postposition label meaning before or in front.
It remains fixed in the released vocabulary pending a narrower named
mapping. \\
851 & \texttt{\textless{}MORPH\_PSP\_MUN\_IRUNTU\textgreater{}} & legacy
postposition & Legacy FST postposition label meaning from before or in
front of. It remains fixed in the released vocabulary pending a narrower
named mapping. \\
852 & \texttt{\textless{}MORPH\_PSP\_NADUVIL\textgreater{}} & legacy
postposition & Legacy FST postposition label meaning in the middle of.
It remains fixed in the released vocabulary pending a narrower named
mapping. \\
853 & \texttt{\textless{}MORPH\_PSP\_NADUVIL\_IRUNTU\textgreater{}} &
legacy postposition & Legacy FST postposition label meaning from in the
middle of. It remains fixed in the released vocabulary pending a
narrower named mapping. \\
854 & \texttt{\textless{}MORPH\_PSP\_PIN\textgreater{}} & legacy
postposition & Legacy FST postposition label meaning after or behind. It
remains fixed in the released vocabulary pending a narrower named
mapping. \\
855 & \texttt{\textless{}MORPH\_PSP\_PINNAAL\textgreater{}} & legacy
postposition & Legacy FST postposition label meaning after or behind. It
remains fixed in the released vocabulary pending a narrower named
mapping. \\
856 & \texttt{\textless{}MORPH\_PSP\_PINNAAL\_IRUNTU\textgreater{}} &
legacy postposition & Legacy FST postposition label meaning from after
or behind. It remains fixed in the released vocabulary pending a
narrower named mapping. \\
857 & \texttt{\textless{}MORPH\_PSP\_PINNE\textgreater{}} & legacy
postposition & Legacy FST postposition label meaning after or behind. It
remains fixed in the released vocabulary pending a narrower named
mapping. \\
858 & \texttt{\textless{}MORPH\_PSP\_PINNE\_IRUNTU\textgreater{}} &
legacy postposition & Legacy FST postposition label meaning from after
or behind. It remains fixed in the released vocabulary pending a
narrower named mapping. \\
859 & \texttt{\textless{}MORPH\_PSP\_PIN\_IRUNTU\textgreater{}} & legacy
postposition & Legacy FST postposition label meaning from after or
behind. It remains fixed in the released vocabulary pending a narrower
named mapping. \\
860 & \texttt{\textless{}MORPH\_PSP\_PIRAKU\textgreater{}} & legacy
postposition & Legacy FST postposition label meaning after. It remains
fixed in the released vocabulary pending a narrower named mapping. \\
861 & \texttt{\textless{}MORPH\_PSP\_POL\textgreater{}} & legacy
postposition & Legacy FST postposition label meaning like or as. It
remains fixed in the released vocabulary pending a narrower named
mapping. \\
862 & \texttt{\textless{}MORPH\_PSP\_POLA\textgreater{}} & legacy
postposition & Legacy FST postposition label meaning like or as. It
remains fixed in the released vocabulary pending a narrower named
mapping. \\
863 & \texttt{\textless{}MORPH\_PSP\_TAVIRA\textgreater{}} & legacy
postposition & Legacy FST postposition label meaning except. It remains
fixed in the released vocabulary pending a narrower named mapping. \\
864 & \texttt{\textless{}MORPH\_PSP\_ULE\textgreater{}} & legacy
postposition & Legacy FST postposition label meaning inside. It remains
fixed in the released vocabulary pending a narrower named mapping. \\
865 & \texttt{\textless{}MORPH\_PSP\_ULE\_IRUNTU\textgreater{}} & legacy
postposition & Legacy FST postposition label meaning from inside. It
remains fixed in the released vocabulary pending a narrower named
mapping. \\
866 & \texttt{\textless{}MORPH\_PSP\_ULLE\textgreater{}} & legacy
postposition & Legacy FST postposition label meaning inside. It remains
fixed in the released vocabulary pending a narrower named mapping. \\
867 & \texttt{\textless{}MORPH\_PSP\_ULLE\_IRUNTU\textgreater{}} &
legacy postposition & Legacy FST postposition label meaning from inside.
It remains fixed in the released vocabulary pending a narrower named
mapping. \\
868 & \texttt{\textless{}MORPH\_PSP\_UL\_IRUNTU\textgreater{}} & legacy
postposition & Legacy FST postposition label meaning from inside. It
remains fixed in the released vocabulary pending a narrower named
mapping. \\
869 & \texttt{\textless{}MORPH\_PSP\_VALIYAAKA\textgreater{}} & legacy
postposition & Legacy FST postposition label meaning through or by way
of. It remains fixed in the released vocabulary pending a narrower named
mapping. \\
870 & \texttt{\textless{}MORPH\_PSP\_VARAIKKUM\textgreater{}} & legacy
postposition & Legacy FST postposition label meaning until or up to. It
remains fixed in the released vocabulary pending a narrower named
mapping. \\
871 & \texttt{\textless{}MORPH\_PSP\_VARAIYIL\textgreater{}} & legacy
postposition & Legacy FST postposition label meaning until or within the
limit. It remains fixed in the released vocabulary pending a narrower
named mapping. \\
872 & \texttt{\textless{}MORPH\_PSP\_VELIYEE\textgreater{}} & legacy
postposition & Legacy FST postposition label meaning outside. It remains
fixed in the released vocabulary pending a narrower named mapping. \\
873 & \texttt{\textless{}MORPH\_PSP\_VELIYEE\_IRUNTU\textgreater{}} &
legacy postposition & Legacy FST postposition label meaning from
outside. It remains fixed in the released vocabulary pending a narrower
named mapping. \\
874 & \texttt{\textless{}MORPH\_PSP\_VELIYIL\textgreater{}} & legacy
postposition & Legacy FST postposition label meaning outside. It remains
fixed in the released vocabulary pending a narrower named mapping. \\
875 & \texttt{\textless{}MORPH\_PSP\_VELIYIL\_IRUNTU\textgreater{}} &
legacy postposition & Legacy FST postposition label meaning from
outside. It remains fixed in the released vocabulary pending a narrower
named mapping. \\
876 & \texttt{\textless{}MORPH\_REGISTER\textgreater{}} & legacy FST
label & Legacy FST label for register label; retained as a fixed
readable factor rather than created dynamically. \\
877 & \texttt{\textless{}MORPH\_SANDHI\_C\textgreater{}} & legacy FST
label & Legacy FST label for legacy c-type sandhi; retained as a fixed
readable factor rather than created dynamically. \\
878 & \texttt{\textless{}MORPH\_SANDHI\_K\textgreater{}} & legacy FST
label & Legacy FST label for legacy k-type sandhi; retained as a fixed
readable factor rather than created dynamically. \\
879 & \texttt{\textless{}MORPH\_SANDHI\_P\textgreater{}} & legacy FST
label & Legacy FST label for legacy p-type sandhi; retained as a fixed
readable factor rather than created dynamically. \\
880 & \texttt{\textless{}MORPH\_SANDHI\_T\textgreater{}} & legacy FST
label & Legacy FST label for legacy t-type sandhi; retained as a fixed
readable factor rather than created dynamically. \\
881 & \texttt{\textless{}MORPH\_UNTIL\textgreater{}} & legacy FST label
& Legacy FST label for until or boundary meaning; retained as a fixed
readable factor rather than created dynamically. \\
882 & \texttt{\textless{}MORPH\_VPARTP\_TAANDI\textgreater{}} & legacy
verbal relation & Legacy participial relation meaning beyond or
crossing. \\
883 & \texttt{\textless{}MORPH\_VPART\_CUTTI\textgreater{}} & legacy
verbal relation & Legacy verbal-participle relation meaning around or
concerning. \\
884 & \texttt{\textless{}MORPH\_VPART\_KONDU\textgreater{}} & legacy
verbal relation & Legacy verbal-participle relation meaning with, by, or
while doing. \\
885 & \texttt{\textless{}MORPH\_VPART\_OTTI\textgreater{}} & legacy
verbal relation & Legacy verbal-participle relation meaning adjoining or
in relation to. \\
886 & \texttt{\textless{}MORPH\_VPART\_TAANDI\textgreater{}} & legacy
verbal relation & Legacy verbal-participle relation meaning beyond or
crossing. \\
887 & \texttt{\textless{}MORPH\_VPART\_TAVIRTU\textgreater{}} & legacy
verbal relation & Legacy verbal-participle relation meaning excluding or
avoiding. \\
888 & \texttt{\textless{}MORPH\_VPART\_VAITTU\textgreater{}} & legacy
verbal relation & Legacy verbal-participle relation meaning using,
keeping, or having done. \\
889 & \texttt{\textless{}MORPH\_VPART\_VIDA\textgreater{}} & legacy
verbal relation & Legacy verbal-participle relation meaning than or
leaving. \\
890 & \texttt{\textless{}NEGATIVE\_PARTICIPLE\textgreater{}} &
participle & Marks a negative non-finite or modifying verb form. \\
891 & \texttt{\textless{}NUM\_CARDINAL\textgreater{}} & number and
quantity & Cardinal number: one, two, three, and so on. \\
892 & \texttt{\textless{}NUM\_FRACTION\textgreater{}} & number and
quantity & Fractional number. \\
893 & \texttt{\textless{}NUM\_ORDINAL\textgreater{}} & number and
quantity & Ordinal number: first, second, and so on. \\
894 & \texttt{\textless{}NUM\_PL\textgreater{}} & number and quantity &
Plural number. \\
895 & \texttt{\textless{}NUM\_SG\textgreater{}} & number and quantity &
Singular number. \\
896 & \texttt{\textless{}PART\_ONLY\textgreater{}} & grammatical or
semantic feature & Restrictive particle: only or just. \\
897 & \texttt{\textless{}PERSON\_1PL\textgreater{}} & person and
agreement & First person plural: we. \\
898 & \texttt{\textless{}PERSON\_1SG\textgreater{}} & person and
agreement & First person singular: I. \\
899 & \texttt{\textless{}PERSON\_2PL\textgreater{}} & person and
agreement & Second person plural: you (plural). \\
900 & \texttt{\textless{}PERSON\_2PL\_HON\textgreater{}} & person and
agreement & Second person plural honorific: respectful you. \\
901 & \texttt{\textless{}PERSON\_2SG\textgreater{}} & person and
agreement & Second person singular: you. \\
902 & \texttt{\textless{}PERSON\_2SG\_HON\textgreater{}} & person and
agreement & Second person singular honorific: respectful you. \\
903 & \texttt{\textless{}PERSON\_3PL\textgreater{}} & person and
agreement & Third person plural: they. \\
904 & \texttt{\textless{}PERSON\_3PL\_EPICENE\textgreater{}} & person
and agreement & Third person plural without a masculine/feminine
distinction. \\
905 & \texttt{\textless{}PERSON\_3PL\_NEUT\textgreater{}} & person and
agreement & Third person plural neuter or non-human. \\
906 & \texttt{\textless{}PERSON\_3SG\textgreater{}} & person and
agreement & Third person singular. \\
907 & \texttt{\textless{}PERSON\_3SG\_EPICENE\textgreater{}} & person
and agreement & Third person singular without a masculine/feminine
distinction. \\
908 & \texttt{\textless{}PERSON\_3SG\_FEM\textgreater{}} & person and
agreement & Third person singular feminine: she. \\
909 & \texttt{\textless{}PERSON\_3SG\_HON\textgreater{}} & person and
agreement & Third person singular honorific. \\
910 & \texttt{\textless{}PERSON\_3SG\_MASC\textgreater{}} & person and
agreement & Third person singular masculine: he. \\
911 & \texttt{\textless{}PERSON\_3SG\_NEUT\textgreater{}} & person and
agreement & Third person singular neuter: it. \\
912 & \texttt{\textless{}POLARITY\_NEG\textgreater{}} & polarity &
Negative polarity. \\
913 & \texttt{\textless{}POLARITY\_POS\textgreater{}} & polarity &
Positive polarity. \\
914 & \texttt{\textless{}POSTPOSITION\textgreater{}} & postposition and
relation & General postposition or relational function word. \\
915 & \texttt{\textless{}POST\_ABOUT\textgreater{}} & postposition and
relation & Relation meaning about or concerning. \\
916 & \texttt{\textless{}POST\_ACCORDING\_TO\textgreater{}} &
postposition and relation & Relation meaning according to or in the
manner stated. \\
917 & \texttt{\textless{}POST\_AFTER\textgreater{}} & postposition and
relation & Temporal or spatial relation meaning after or behind. \\
918 & \texttt{\textless{}POST\_AMONG\textgreater{}} & postposition and
relation & Relation meaning among or between. \\
919 & \texttt{\textless{}POST\_BEFORE\textgreater{}} & postposition and
relation & Temporal or spatial relation meaning before or in front
of. \\
920 & \texttt{\textless{}POST\_LIKE\_AS\textgreater{}} & postposition
and relation & Similarity relation: like or as. \\
921 & \texttt{\textless{}POST\_TOWARD\textgreater{}} & postposition and
relation & Direction relation: toward. \\
922 & \texttt{\textless{}POST\_UNTIL\textgreater{}} & postposition and
relation & Boundary relation: until or up to. \\
923 & \texttt{\textless{}POST\_WITHIN\_BY\textgreater{}} & postposition
and relation & Interior or deadline relation: within, inside, or by. \\
924 & \texttt{\textless{}POS\_ADJ\textgreater{}} & part of speech & Part
of speech: adjective. \\
925 & \texttt{\textless{}POS\_ADV\textgreater{}} & part of speech & Part
of speech: adverb. \\
926 & \texttt{\textless{}POS\_INTERJECTION\textgreater{}} & part of
speech & Part of speech: interjection. \\
927 & \texttt{\textless{}POS\_NOUN\textgreater{}} & part of speech &
Part of speech: noun. \\
928 & \texttt{\textless{}POS\_PART\textgreater{}} & part of speech &
Part of speech: particle. \\
929 & \texttt{\textless{}POS\_PARTICIPIAL\_NOUN\textgreater{}} & part of
speech & Part of speech: noun formed from a participle. \\
930 & \texttt{\textless{}POS\_PRONOUN\textgreater{}} & part of speech &
Part of speech: pronoun. \\
931 & \texttt{\textless{}POS\_QUANTIFIER\textgreater{}} & part of speech
& Part of speech: quantifier. \\
932 & \texttt{\textless{}POS\_VERBAL\_NOUN\textgreater{}} & part of
speech & Part of speech: verb-derived action or event noun. \\
933 & \texttt{\textless{}PRESENTATIVE\textgreater{}} & grammatical or
semantic feature & Presentative expression used to point out or
introduce something. \\
934 & \texttt{\textless{}PRIVATIVE\_WITHOUT\textgreater{}} & grammatical
or semantic feature & Privative meaning: without or lacking. \\
935 & \texttt{\textless{}PRON\_EXCLUSIVE\textgreater{}} & pronoun &
Exclusive first-person plural: we, excluding the addressee. \\
936 & \texttt{\textless{}PRON\_INCLUSIVE\textgreater{}} & pronoun &
Inclusive first-person plural: we, including the addressee. \\
937 & \texttt{\textless{}PRON\_POSSESSIVE\textgreater{}} & pronoun &
Possessive pronoun function. \\
938 & \texttt{\textless{}PRON\_REFLEXIVE\textgreater{}} & pronoun &
Reflexive pronoun function: self. \\
939 & \texttt{\textless{}QUANT\_ALL\textgreater{}} & grammatical or
semantic feature & Universal quantity: all or every. \\
940 & \texttt{\textless{}RECIPROCAL\textgreater{}} & grammatical or
semantic feature & Reciprocal relation: each other. \\
941 & \texttt{\textless{}REDUPLICATION\textgreater{}} & grammatical or
semantic feature & Marks a repeated form used for distribution,
emphasis, or iteration. \\
942 & \texttt{\textless{}REGISTER\_COLLOQUIAL\textgreater{}} &
grammatical or semantic feature & Marks a colloquial form. \\
943 & \texttt{\textless{}REL\_ATTACH\textgreater{}} & grammatical or
semantic feature & Marks an attaching or related-to construction. \\
944 & \texttt{\textless{}SANDHI\_C\textgreater{}} & sandhi & Marks
c-type linking sandhi. \\
945 & \texttt{\textless{}SANDHI\_K\textgreater{}} & sandhi & Marks
k-type linking sandhi. \\
946 & \texttt{\textless{}SANDHI\_P\textgreater{}} & sandhi & Marks
p-type linking sandhi. \\
947 & \texttt{\textless{}SANDHI\_T\textgreater{}} & sandhi & Marks
t-type linking sandhi. \\
948 & \texttt{\textless{}SEM\_HUMAN\textgreater{}} & grammatical or
semantic feature & Marks reference to a human being or group. \\
949 & \texttt{\textless{}SEM\_PURPOSE\textgreater{}} & grammatical or
semantic feature & Marks purpose or intended use. \\
950 & \texttt{\textless{}STEM\_OBLIQUE\textgreater{}} & grammatical or
semantic feature & Marks a changed noun stem used before a case
ending. \\
951 & \texttt{\textless{}TEMPORAL\_IMMEDIATE\textgreater{}} &
grammatical or semantic feature & Marks immediate succession: as soon
as. \\
952 & \texttt{\textless{}TEMPORAL\_WHEN\textgreater{}} & grammatical or
semantic feature & Marks a time relation: when or while. \\
953 & \texttt{\textless{}TENSE\_FUTURE\textgreater{}} & tense & Future
tense. \\
954 & \texttt{\textless{}TENSE\_PAST\textgreater{}} & tense & Past
tense. \\
955 & \texttt{\textless{}TENSE\_PRESENT\textgreater{}} & tense & Present
tense. \\
956 & \texttt{\textless{}TITLE\_HONORIFIC\textgreater{}} & grammatical
or semantic feature & Marks an honorific title. \\
957 & \texttt{\textless{}VERBAL\_PARTICIPLE\textgreater{}} & verb form &
Marks a non-finite verb that links to a following action. \\
958 & \texttt{\textless{}VERB\_COMPLEX\textgreater{}} & verb form &
Broad inherited FST label for a complex verb construction. \\
959 & \texttt{\textless{}VERB\_FINITE\textgreater{}} & verb form & Broad
inherited FST label for a finite verb. \\
960 & \texttt{\textless{}VERB\_IMPERATIVE\textgreater{}} & verb form &
Imperative verb form: a command or request. \\
961 & \texttt{\textless{}VERB\_INFINITIVE\textgreater{}} & verb form &
Infinitive verb form. \\
962 & \texttt{\textless{}VERB\_NONFINITE\textgreater{}} & verb form &
Broad inherited FST label for a non-finite verb. \\
963 & \texttt{\textless{}VOICE\_CAUSATIVE\textgreater{}} & voice &
Causative voice: causes someone or something to act. \\
964 & \texttt{\textless{}VOICE\_PASSIVE\textgreater{}} & voice & Passive
voice: presents the affected participant rather than the actor. \\
\end{longtable}
}

\endgroup

\end{landscape}\clearpage

\section{References}\label{references}

\begin{itemize}
\tightlist
\item
  Doddapaneni, S., et al. 2023.
  \href{https://aclanthology.org/2023.acl-long.693/}{Towards Leaving No
  Indic Language Behind: Building Monolingual Corpora, Benchmark and
  Models for Indic Languages}. ACL.
\item
  Gala, J., et al. 2023.
  \href{https://openreview.net/forum?id=vfT4YuzAYA}{IndicTrans2: Towards
  High-Quality and Accessible Machine Translation Models for all 22
  Scheduled Indian Languages}. \emph{Transactions on Machine Learning
  Research}.
\item
  Khan, M. S. U. R., et al. 2024.
  \href{https://arxiv.org/abs/2403.06350}{IndicLLMSuite: A Blueprint for
  Creating Pre-training and Fine-Tuning Datasets for Indian Languages}.
\item
  Kudo, T., and Richardson, J. 2018.
  \href{https://aclanthology.org/D18-2012/}{SentencePiece: A Simple and
  Language Independent Subword Tokenizer and Detokenizer for Neural Text
  Processing}. EMNLP System Demonstrations.
\item
  Merity, S., Xiong, C., Bradbury, J., and Socher, R. 2017.
  \href{https://openreview.net/forum?id=Byj72udxe}{Pointer Sentinel
  Mixture Models}. ICLR.
\item
  Mhaske, A., et al. 2023.
  \href{https://aclanthology.org/2023.acl-long.582/}{Naamapadam: A
  Large-Scale Named Entity Annotated Data for Indic Languages}. ACL.
\item
  NLLB Team, et al. 2022. \href{https://arxiv.org/abs/2207.04672}{No
  Language Left Behind: Scaling Human-Centered Machine Translation}.
\item
  Popović, M. 2017. \href{https://aclanthology.org/W17-4770/}{chrF++:
  Words Helping Character N-grams}. WMT.
\item
  Press Information Bureau, Government of India.
  \href{https://www.pib.gov.in/}{Press releases} and
  \href{https://www.pib.gov.in/Content/102_2_Copyright-Policy.aspx?lang=1&reg=3}{copyright
  policy}.
\item
  Ramasamy, L., and Žabokrtský, Z. 2012.
  \href{http://www.lrec-conf.org/proceedings/lrec2012/summaries/456.html}{Prague
  Dependency Style Treebank for Tamil}. LREC.
\item
  Ramesh, G., et al. 2022.
  \href{https://aclanthology.org/2022.tacl-1.9/}{Samanantar: The Largest
  Publicly Available Parallel Corpora Collection for 11 Indic
  Languages}. \emph{Transactions of the Association for Computational
  Linguistics}, 10, 145--162.
\item
  Rei, R., et al. 2022.
  \href{https://aclanthology.org/2022.wmt-1.60/}{CometKiwi: IST-Unbabel
  2022 Submission for the Quality Estimation Shared Task}. WMT.
\item
  Sarveswaran, K., Dias, G., and Butt, M. 2019.
  \href{https://aclanthology.org/W19-3111/}{Using Meta-Morph Rules to
  Develop Morphological Analysers: A Case Study Concerning Tamil}.
  FSMNLP.
\item
  Sarveswaran, K., Dias, G., and Butt, M. 2021.
  \href{https://doi.org/10.1007/s10590-021-09261-5}{ThamizhiMorph: A
  Morphological Parser for the Tamil Language}. \emph{Machine
  Translation}, 35(1), 37--70.
\item
  Sennrich, R., Haddow, B., and Birch, A. 2016.
  \href{https://aclanthology.org/P16-1162/}{Neural Machine Translation
  of Rare Words with Subword Units}. ACL.
\item
  Sutton, R. S. 2019.
  \href{http://www.incompleteideas.net/IncIdeas/BitterLesson.html}{The
  Bitter Lesson}.
\item
  Su, J., et al. 2021. \href{https://arxiv.org/abs/2104.09864}{RoFormer:
  Enhanced Transformer with Rotary Position Embedding}.
\item
  University of Madras. 1924--1936.
  \href{https://dsal.uchicago.edu/dictionaries/tamil-lex/}{\emph{Tamil
  Lexicon}}. University of Madras; digital database hosted by the
  University of Chicago Digital South Asia Library.
\item
  Vaswani, A., et al. 2017.
  \href{https://arxiv.org/abs/1706.03762}{Attention Is All You Need}.
\item
  Vuizur.
  \href{https://github.com/Vuizur/Wiktionary-Dictionaries}{\emph{Wiktionary-Dictionaries}}.
  Supplemental Tamil--English data extracted from Wiktionary.
\item
  Wiktionary contributors.
  \href{https://ta.wiktionary.org/wiki/முதற்_பக்கம்}{Tamil Wiktionary}.
  Wikimedia Foundation; source snapshots obtained from the official
  database dumps.
\end{itemize}

\end{document}